\documentclass[12pt]{article}

\usepackage{amssymb,amsmath,amsfonts,eurosym,geometry,ulem,graphicx,caption,color,setspace,sectsty,comment,footmisc,caption,natbib,pdflscape,subfigure,array,hyperref}

\newcolumntype{L}[1]{>{\raggedright\let\newline\\arraybackslash\hspace{0pt}}m{#1}}
\newcolumntype{C}[1]{>{\centering\let\newline\\arraybackslash\hspace{0pt}}m{#1}}
\newcolumntype{R}[1]{>{\raggedleft\let\newline\\arraybackslash\hspace{0pt}}m{#1}}

\usepackage{endnotes}
\let\footnote=\endnote

\usepackage{graphicx}
\usepackage{multirow}
\usepackage{hhline}

\usepackage{epsfig}    
\usepackage{subfigure} 
\usepackage{comment}
\usepackage{mathrsfs}
\usepackage{amssymb}
\usepackage{url}
\usepackage{amsmath}
\usepackage{rotating} 
\usepackage{multirow,makecell}
\usepackage{threeparttable}
\usepackage{booktabs}
\usepackage{algorithmicx}
\usepackage[ruled]{algorithm}
\usepackage{algpseudocode}
\usepackage{color}
\usepackage{bm}
\usepackage{graphicx}
\usepackage{caption}
\usepackage{authblk}

\newcommand\figcaption{\def\@captype{figure}\caption}
\newcommand\tabcaption{\def\@captype{table}\caption}

\usepackage{appendix}
\usepackage{color}
\definecolor{strcolor}{rgb}{0.6, 0.2, 0.6}
\definecolor{commentcolor}{rgb}{0.3125, 0.5, 0.3125}
\definecolor{keycol}{rgb}{0, 0, 1}

\usepackage{bbm}

\usepackage{listings}
\usepackage{url}

\newcommand {\bea}{\begin{eqnarray}}
	\newcommand {\eea}{\end{eqnarray}}

\def\blot{\quad \mbox{$\vcenter{ \vbox{ \hrule height.4pt
				\hbox{\vrule width.4pt height.9ex \kern.9ex \vrule width.4pt}
				\hrule height.4pt}}$}}

\usepackage{natbib}
\bibpunct[, ]{(}{)}{,}{a}{}{,}%
\usepackage{setspace}
\newtheorem{definition}{Definition}[section]  

\begin{document}

\title{Modeling Reference-dependent Choices with Graph Neural Networks}
\author[1]{Liang Zhang}
\author[2]{Guannan Liu}
\author[2]{Junjie Wu}
\author[3]{Yong Tan}
\affil[1]{College of Computing and Data Science, Nanyang Technological University}
\affil[2]{School of Economics and Management, Beihang University}
\affil[3]{Foster School of Business, University of Washington}

\date{\today}
\maketitle
\begin{abstract}
\noindent Understanding consumer behaviors and preferences is paramount for developing effective recommender systems and enhancing business profitability. While the classic Prospect Theory has highlighted the reference-dependent and comparative nature of consumers’ product evaluation processes, few models have successfully integrated this theoretical hypothesis into data-driven preference quantification, particularly in the realm of recommender systems development. To bridge this gap, we propose a new research problem of modeling reference-dependent preferences from a data-driven perspective, and design a novel deep learning-based framework named \underline{A}ttributed \underline{R}eference-dependent \underline{C}hoice Model for \underline{Rec}ommendation (\emph{ArcRec}) to tackle the inherent challenges associated with this problem. \emph{ArcRec} features in building a reference network from aggregated historical purchase records for instantiating theoretical reference points, which is then decomposed into product attribute specific sub-networks and represented through Graph Neural Networks for fine-grained referencing. In this way, the reference points of a consumer can be encoded at the attribute-level individually from her past experiences but also reflect the crowd influences.  \emph{ArcRec} also makes novel contributions to  quantifying consumers' reference-dependent preferences using a deep neural network-based utility function that integrates both interest-inspired and price-inspired preferences, with their complex interaction effects captured by an attribute-aware price sensitivity mechanism. Most importantly, \emph{ArcRec} introduces a novel Attribute-level Willingness-To-Pay measure to the reference-dependent utility function, which captures a consumer's heterogeneous salience of product attributes via observing her attribute-level price tolerance to a product. Empirical evaluations on both synthetic and real-world online shopping datasets demonstrate \emph{ArcRec}'s superior performances over fourteen state-of-the-art baselines, particularly in scenarios involving cold-start products with limited consumption history. A controlled experiment simulating price changes and several illustrative cases further validate \emph{ArcRec}'s ability in effectively capturing consumers' heterogeneous price sensitivity and identifying their personalized salience of product attributes, thereby leading to better recommendation results. \\

\end{abstract}
	
\section{Introduction}
Modeling consumer choices has been a longstanding challenge to understand consumer behaviors and bears significant implications for the design of better marketing and promotional strategies to boost sales and profits~\citep{fader1996modeling,andrews1998simulation}. In the literature, choice models have been extensively studied in marketing and management science fields, with a primary focus on the comparison and trade-off among product alternatives~\citep{kamakura1989probabilistic}. 
As posited by the classic Prospect Theory~\citep{tversky1991loss}, consumers' evaluations of products are reference-dependent and comparative in nature \citep{mussweiler2003comparison, kahneman2013prospect}. The  decision-making process of consumers, according to the theory, involves comparing a target product against reference points. Along this line, tremendous research efforts have been dedicated to validating the effects of reference points on consumer preferences and choices, most of which have predominantly relied on theoretical analysis or empirical findings~\citep{dhar1999comparison, dhar1992effect} to demonstrate the importance of reference points modeling for accurate reflection of consumer preferences and purchasing decisions across various contexts~\citep{arkes2008reference, monroe2015examining, schwartz2008prospect}.

With the proliferation of \emph{e}-commerce platforms and the recent advances in data-driven models like deep neural networks, an abundance of recommendation methods have emerged, focusing on modeling consumers' choices and preferences from their historical consumption data \citep{bell2007improved,rendle2012bpr,xue2019deep,li2023next,zhang2024denoising, wang2024efficient}. It is reported that high-quality recommendations have a significant impact on consumers’ purchase decisions \citep{pathak2010empirical}. For example, 60\% of Netflix rentals and 35\% of Amazon’s sales are attributed to their recommendation systems \citep{hosanagar2014will}.
The general idea of recommender systems is to estimate personalized product utilities for consumers \citep{xiao2007commerce}, which essentially aligns with the construction of product utility among various alternatives in choice models. However, while the Prospect Theory and reference points have been proven beneficial to modeling consumer choices by substantial empirical evidence, this theoretical foundation is largely overlooked by existing recommendation methods, leading to potentially less satisfactory recommender systems. This indeed motivates our pilot study in this paper on developing a data-driven reference-dependent choice model for substantial improvement of product recommendations. This can be also regarded as a beneficial trial in bridging the vast gap between managerial theories and advanced machine learning methods.


The research task mentioned above is indeed non-trivial and poses several critical challenges. First, reference points were theoretically considered as consumers' subjective expectations and associated knowledge derived from their past experiences \citep{winer1986reference, bordalo2013salience}, and therefore, can hardly be instantiated in a data-driven manner from the traditional standpoint. Several attempts have been made to address this by utilizing the last purchased product as the reference point, implying that reference points are formed solely based on a consumer's recent purchases~\citep{hardie1993modeling,briesch1997comparative}.
Nevertheless, it is intuitive that when making purchase decision on a target product, consumers often refer to not only their own purchase experiences but also the aggregated preferences of the crowd. For instance, the ``also-viewed'' or ``also-purchased'' recommendation links generated from the crowd’s collective purchase records in \emph{e}-commerce sites~\citep{dhar2014prediction,lin2017misq} can significantly influence consumers' shopping decisions and serve as additional sources of their reference points. Hence, it is necessary to integrate both personalized experiences and the \emph{wisdom of crowds} to instantiate reference points from purchase data. 

Moreover, it is often deemed coarse-grained to construct consumers' reference-dependent preferences at the product-level~\citep{andrews1999mds}, by comparing a target product with referenced products through a complex information processing process. Instead, comparisons at the product attribute-level, \emph{e.g.}, two cell-phones with different brands or functionalities, are believed to be more sensible~\citep{fader1996modeling}. Several empirical findings have also demonstrated that consumers often use distinct reference points for different product attributes~\citep{wang2019incorporating}. However, how to model reference points at the fine-grained attribute level in a data-driven manner, remains a crucial challenge. As a natural extension to the above argument, the reference points identified with respect to different product attributes might contribute differently to the perceived utilities due to consumers' heterogeneous attribute salience --- a key factor advocated by the Multiple Attribute Utility Theory~\citep{keeney1993decisions} in modeling consumers' choices. For instance, some consumers may prioritize brand over material and style when choosing a handbag; in consequence, their decisions are more seriously impacted by brand-specific reference points. Despite its importance, to our best knowledge, few efforts have been devoted to modeling the personalized salience of product attributes from consumers' purchase data.

Additionally, price, as a particular type of product attribute, is often specially considered in modeling reference-dependent preferences, known as \emph{reference price}~\citep{lattin1989reference, winer1986reference}. It has been empirically demonstrated that reference price plays a fundamental role in understanding potential effects of reference points on consumer choices~\citep{mazumdar2005reference}, and was generally modeled at the product-level~\citep{lattin1989reference, chang1999impact}.
However, it has been found that product price often interacts extensively with other product attributes to shape consumers' reference-dependent preferences~\citep{chen2014does}. For example, when purchasing a handbag, consumers may evaluate whether the brand is worth the price compared with their expectations, and those with a strong interest for a specific brand could even accept a premium. Therefore, modeling the special impact of product price on consumers' choices with respect to identified attribute-level reference points in a data-driven manner, is an intriguing and open question that calls for further investigation.

To address the above-mentioned challenges, we propose a novel deep learning based recommendation framework named \underline{A}ttributed \underline{R}eference-dependent \underline{C}hoice Model for \underline{Rec}ommendation (\emph{ArcRec}) in this paper. \emph{ArcRec} generally features in designing a data-driven, attribute-level reference-dependent choice model that integrates both the interest-inspired and price-inspired preferences of consumers. Specifically, we first construct a novel \emph{reference network} from aggregated historical purchase records, with the purpose of capturing all consumers’ collective purchase behaviors, wherein each product is represented as a node and bi-directional edges are established between every pair of co-purchased products in the recent past. From this reference network, a consumer can form her reference points from the perspectives of both her own and the entire crowd’s past purchases through high-order message propagation  along the graph. 
To accommodate fine-grained modeling of attribute-level reference points, we further decompose the reference network into various \emph{attributed reference networks} (ARN) corresponding to different product attributes. Through analyzing the structures of these ARNs by Graph Neural Networks (GNNs), we can obtain product embeddings from a consumer's past purchases, which serve as personalized and diverse reference points at the fine-grained attribute level. We then measure a consumer's reference-dependent preference by comparing a target product against reference points, which is decomposed into interest-inspired and price-inspired preferences to account for the two major gauges in consumer choices. Particularly, for capturing the price-inspired preference, we introduce a novel \emph{attribute-aware price sensitivity} mechanism to model the interaction effects between product price and attributes. All these components are expressed using powerful deep neural networks to simulate the complex information processing process involved in decision-making. Moreover, to integrate the reference-dependent utilities driven by different attribute-level reference points, we define a novel \emph{attribute-level willingness-to-pay} (AWTP) measure that evaluates the tolerance of a consumer to the extra price volatility given a product attribute, aiming to capture consumers' heterogeneous salience of product attributes from their past purchases. Finally, we integrate the proposed deep learning based reference-dependent choice model into the Bayesian Personalized Ranking framework~\citep{rendle2012bpr} for end-to-end model training, from which recommendations can be generated with the learned parameters.

Empirical evaluations on both synthetic and a real-world online shopping datasets have been implemented and demonstrated the superior performances of \emph{ArcRec} over the state-of-the-art recommendation methods. In particular, \emph{ArcRec} exhibits remarkable performances on cold-start or new products with limited historical purchase records, highlighting the precious value of attributed reference networks in addressing this longstanding and intractable challenge. In addition, to validate  \emph{ArcRec}'s ability in capturing consumers' heterogeneous price sensitivity in utility perception, a controlled experiment is conducted by simulating the price change and monitoring the resulting changes in product rankings for different consumers. The results suggest that the recommendation ranking list generated by \emph{ArcRec} is more significantly altered for price-sensitive consumers; conversely, the rankings show less variations for consumers with lower price sensitivity. These results indicate that our method successfully differentiates consumers' heterogeneous price sensitivity and integrates their price perceptions effectively. Illustrative cases for the AWTP measure are also presented to gain insights into how \emph{ArcRec} can restore consumers' personalized heterogeneous salience of product attributes, demonstrating \emph{ArcRec}'s capability of locating more important attributes for shaping consumers' reference-dependent decisions, thereby enhancing the personalized recommendation experience. 
	
\section{Literature Review}
We draw upon three key bodies of literature that are closely related to our research. The first line of research investigates the modeling of consumer choices with a special focus on reference effects. Secondly, we examine the recent progress in recommendation methods that take product attributes and/or prices into consideration. Finally, we explore the recent advances in deep learning based recommendation methods as they closely relate to our modeling framework. 

\subsection{Reference Effect for Consumer Choices Modeling}
Prospect Theory~\citep{kai1979prospect} has emerged as a seminal contribution in elucidating the human decision-making process. Building upon the theory, consumers are posited to evaluate product choice with respect to a reference point. In particular, \cite{tversky1991loss} develop a theoretical model to capture the effects of reference points on consumer choices. \cite{kHoszegi2006model} consider the reference points as the  expectations about the outcomes given the recent past experiences, which can further guide the learning of reference-dependent preferences.

However, the classic Prospect Theory primarily focuses on choices pertaining to one single attribute. In order to address this limitation, \cite{tversky1991loss} further extend the framework by incorporating multiple attributes into the analysis. Specifically, consumers' choices for a product can be decomposed into a set of values on the product attributes and their corresponding weights~\citep{keeney1993decisions}, thereby a number of research efforts have been proposed to investigate consumers' preferences at the attribute-level. It is generally accepted that consumers tend to form their overall preferences for choice alternatives by evaluating the attributes describing each product, rather than considering the product as a whole~\citep{andrews1999mds}. Meanwhile, it has also been demonstrated that the value of attributes is not solely based on the absolute levels, but rather on its deviation from the reference standard~\citep{kHoszegi2006model,bordalo2013salience}. Additionally, \cite{KOCHER201924} have revealed that the attribute-level anchoring effect can bias consumer choices towards numerical attributes in product recommendations.

As one of the most salient attributes, product price often plays a vital role in shaping consumers' choices. Especially, consumers' perceived value for a product may deviate from its actual price~\citep{sweeney2001consumer}. Therefore, echoing with the Prospect Theory, consumers perceive the price according to a reference price, which has been extensively studied to capture the nuances of consumers' purchasing decisions~\citep{kalyanaram1995empirical}. For instance, \cite{putler1992incorporating} incorporates reference price into the consumer choice models and further investigates the formation of reference effects on consumer behaviors. Analytical evidence has also demonstrated that reference price can significantly impact consumer choice behaviors~\citep{wang2018prospect}. In response to this, various models have been developed to incorporate reference price in modeling consumers' choices~\citep{wang2021reference}. For example, \cite{fibich2005dynamics} derive a mathematical formulation for the price elasticity of demand in the presence of reference price effects. 

Despite the rich research efforts in this line, most of them have predominantly relied on theoretical analysis or empirical findings, lacking the ability to operationalize the estimates for consumers' fine-grained preferences for attributes, price as well as their interplay in a data-driven manner. Our research attempts to bridge this gap by proposing a data-driven approach that utilizes consumers' purchasing records to model their nuanced reference-dependent preferences with regards to product attributes and prices.

\subsection{Recommendation Approaches with Attribute and Price Awareness}
\subsubsection{Attribute-aware Recommendation.}
As demonstrated in prior work, modeling consumers' choices at the granularity of product attributes can effectively capture consumers' fine-grained preferences, leading to improved recommendation performances. Therefore, many researchers have attempted to leverage product attributes to design better recommendation approaches. Indeed, attribute-aware recommendation can simply be regarded as a type of content-based recommendation~\citep{pazzani2007content} when the attributes are treated as contents to derive similarity measures. For example, \cite{manouselis2007experimental} propose multi-attribute collaborative filtering algorithms with emphasis on similarity measurements and feature weighting. In addition, the Factorization Machine (FM) \citep{rendle2010factorization} has become one of the seminal approaches to model the interaction of product attributes and is shown to significantly improve recommendation performances. Furthermore, with the recent success of deep learning, deep variants of FM have also been proposed to enhance the representation capacity of recommendation models. Examples include AFM~\citep{xiao2017afm}, DeepFM~\citep{guo2017deepfm}, PNN~\citep{qu2016pnn}, \emph{etc.}, which model the interactions of attributes with attention mechanisms and deep neural networks such as the simple Multi-Layer Perceptrons (MLP). As another recent research trend, product attributes are also utilized through knowledge graphs, where each attribute is considered as a node connecting to a product. In this regard, a research line aims to use knowledge embeddings to improve the quality of product representations. For example, CKE \citep{zhang2016collaborative} and DKN \citep{wang2018dkn} regard semantic representations learned from knowledge graphs as the extra content information of products. Another research line unifies knowledge graph completion problem with recommendation task. Taking advantages of information propagation proposed by graph neural networks, examples like KGAT \citep{wang2019kgat} and KGNN-LS \citep{wang2019knowledge} perform embedding propagation over attribute knowledge graphs and transform product recommendation to graph link prediction.

In general, while a variety of attribute-based recommendation methods have been proposed, there still lacks a systematic framework to model consumers' heterogeneous and fine-grained attribute-level preferences with theoretical guidance. Our research endeavors to follow the reference effect theories and design a better reference-dependent schema for characterizing the role of different attributes in decision-makings.

\subsubsection{Price-aware Recommendation.}
In the realm of recommendation approaches, there is a specific focus on capturing consumers' preferences towards price beyond common product attributes. Numerous studies have been conducted to design recommendation methods that incorporate price as a significant attribute. These methods often discretize price into categorical features and take into account consumers' price preferences in a personalized manner.
For example, the research by \cite{chen2014does} shows that product price information and consumer price perception could be used to improve the recommendation accuracy in unexplored product categories. \cite{wan2017modeling} model the price sensitivity and preferences from transaction logs with a latent factor model to develop recommendation methods. 
In recent studies by \cite{umberto2015developing} and \cite{zheng2020price}, price-aware recommender systems are developed; however, the product price is processed in a discrete way rather than in a numerical form and the personalization as well as price sensitivity issues are not fully addressed. \cite{ban2021personalized} assume a personalized demand model and dynamically adjust the price of a product in the individual consumer level, by utilizing information about consumers' characteristics. \cite{zhang2022price} model consumers' price preferences in the session recommendation contexts.

While price has been studied from different perspectives, the existing models generally overlook the effect of reference price as well as the pervasive nature of the interactions between price and other attributes in consumer preference modeling. Additionally, most of these methods treat price as an auxiliary attribute to characterize the products, but rarely utilize it to understand how consumers perceive the product value and how it ultimately influences their decision-makings.

\subsection{Deep Learning based Recommendation}
The most classical paradigm of recommender systems is to parameterize consumers and products as vector representations and predict consumer-product interactions based on inner dot~\citep{koren2009matrix, hu2008collaborative}. Though commonly used, the linearity of inner dot makes it insufficient to capture the complex relationships between consumers and products~\citep{liu2021beyond}. Recent years have witnessed great success in exploiting deep learning techniques to enhance recommendations. Along this line, NeuMF \citep{he2017neural} and NNCF \citep{bai2017neural} are proposed to map consumers and products into the embedding space and further employ feedforward neural networks as the interaction function.

Apart from introducing nonlinear interaction functions, the consumer-product graph constructed from historical purchasing records is also exploited to improve recommendation effectiveness. In this regard, Graph Neural Networks (GNN) \citep{welling2016semi, velivckovic2017graph} shed light on capturing the nuances of fine-grained consumer preferences by modeling high-order consumer-product interactions. The methods based on GNNs typically perform information propagation and local neighborhood aggregation to refine consumer and product representations and mainly differ in various aggregation operations~\citep{wang2019neural, wang2020disentangled, he2020lightgcn, chen2020revisiting, kong2022linear}, among which DGCF \citep{wang2020disentangled} and LightGCN \citep{he2020lightgcn} have gained great attentions and are the state-of-the-art methods in this stream. 

Our work is closely related to deep learning-based recommendation, in which we adopt deep learning and graph neural network to model consumers' choices from purchasing records. However, our work stands out in evaluating the personalized utility for a product in a reference-dependent manner and with attribute and price awareness.

\section{Research Motivation and Problem Formulation}
\label{sect:decomp}



Prospect Theory \citep{tversky1991loss} postulates that consumers form judgments based on the evaluation of outcomes relative to reference points. \cite{kHoszegi2006model} further posit that reference points are given by consumers' expectations from their past experiences. Indeed, the importance of reference point has been supported by substantial empirical evidence in both economics and marketing literature.  For example, \cite{hardie1993modeling} find that consumers' most recently purchased brand can shape their expectations, significantly affecting subsequent brand choices. \cite{baucells2010predicting} also unveil that the utility of current consumption is affected by past consumption, known as \emph{reference-dependence}. Building upon the above theory and empirical findings, it is rational to expect that the incorporation of reference points and reference-dependent preferences can offer a more accurate representation of consumers' decision-making behaviors in product choices. 



Despite the great value of reference points for consumer understanding, it is surprising that few existing studies have explicitly modeled reference points in a data-driven manner. This might be ascribed to the fact that reference points are often predominantly shaped by consumers' subjective expectations and associated knowledge derived from their past experiences~\citep{bordalo2013salience, kHoszegi2006model}. As evidenced by~\cite{abeler2011reference}, such information is hard to be observed directly from field data, making it difficult to instantiate reference points for consumer utility evaluation. 

In this paper, we aim to design a data-driven approach to model consumers' reference-dependent preferences for high-performance product recommendation. 
Inspired by the above literature, it is intuitive to assume that a consumer's recently purchased products, as part of her past experiences, can serve as her reference points for making purchase decisions on a focal product. However, it is also generally understood that except for own experiences, a consumer's purchase decision could be greatly influenced by the crowds. For example, the ``also-viewed" or ``also-purchased" recommendation links generated from the crowd's collective purchase records in $e$-commerce sites \citep{dhar2014prediction,lin2017misq}, can also guide a consumer's shopping decisions and serve as another source of her reference points. Hence, it is reasonable to assume that reference points contain influences from both personalized purchased products and the crowd's preferential information, posing challenges to designing appropriate models for reference points quantification.

To meet the above challenge, we first construct a product network from aggregated historical purchase records, with the purpose of capturing all consumers' collective purchase behaviors. In this network, each product is represented as a node, and bi-directional edges are established between every pair of products that were co-purchased by at least one consumer in the recent past.  This product network is termed as the \emph{Reference Network} and formally defined as follows.





\begin{definition}[Reference Network]
A reference network is denoted as $\mathcal{G}=<\mathcal{V},\mathcal{E}>$, where each node $v_i\in \mathcal{V}$ represents a product, and an edge $e_{ij} \in \mathcal{E}$ between nodes $v_i$ and $v_j$ is established if the two products were ever co-purchased by at least one consumer in the observed window period.\hfill$\blacksquare$
\end{definition}

\begin{figure}[t!]
	\centering
	\includegraphics[width=0.9\textwidth]{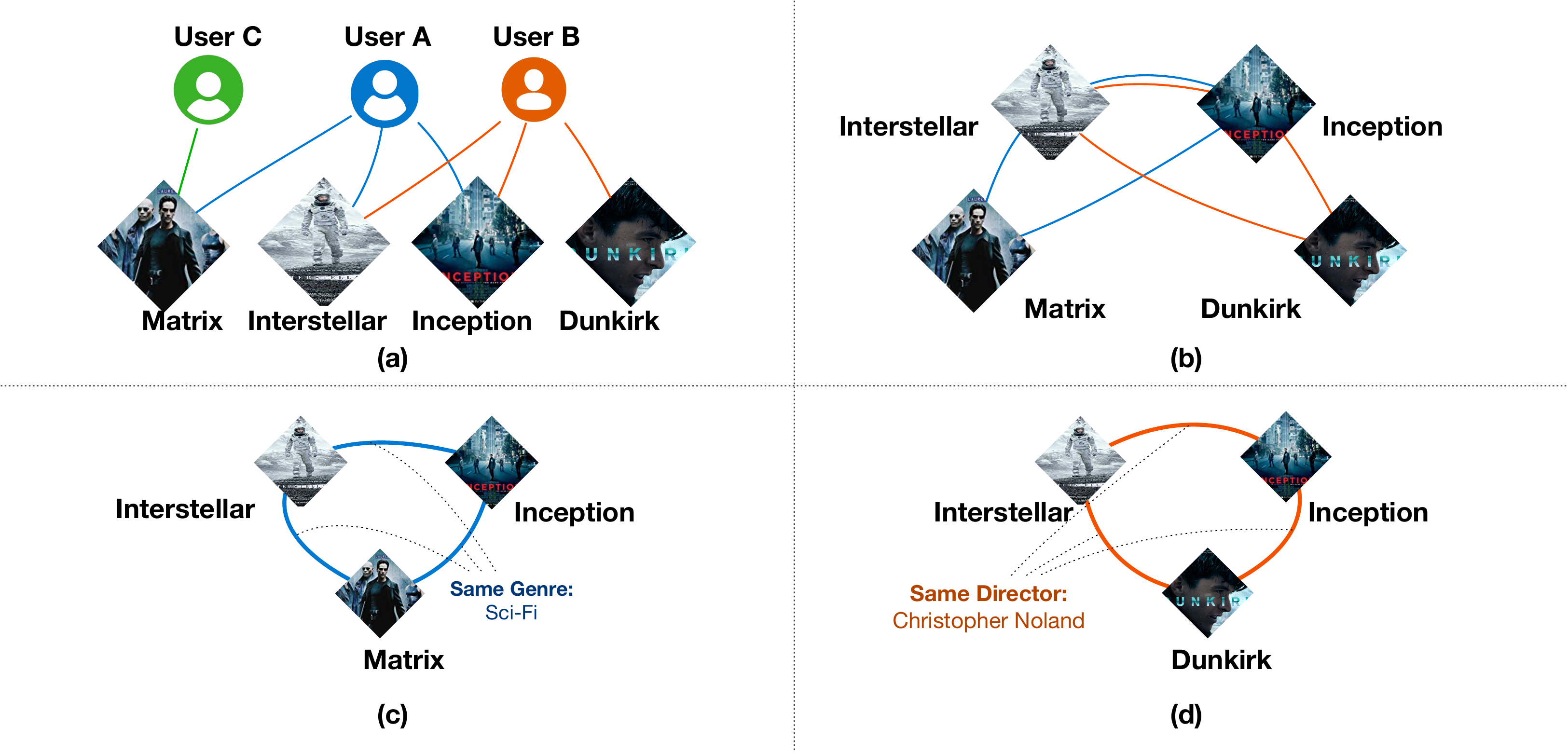}
	\caption{Example of Reference Network and Attributed Reference Network. (a) users' movie watching history; (b) reference network constructed from the movie watching history; (c) genre-specific reference network; (d) director-specific reference network.}
	\label{fig:rdpnex}
\end{figure}

The reference network offers a global view of collective purchase behaviors, from which a consumer can form her reference points from both her own and the entire crowd's past purchases through high-order message propagation in the network. We illustrate this with an example of movie watching as follows. As depicted in Figure~\ref{fig:rdpnex}(a), we can construct a reference network from three users' movie watching records, by treating each movie as a node and establishing edges between those that have been co-watched by at least one user, as shown in Figure~\ref{fig:rdpnex}(b). Notably, the reference network reflects the ``wisdom of crowds'' in purchases, which can contribute to the modeling of a consumer's reference points besides her own behaviors. For instance, consider user C who has only watched the movie \emph{Matrix}. The intuitive modeling of reference points relies solely on this single movie. However, owing to the collective behaviors captured by the reference network, the movies \emph{Interstellar} and \emph{Inception} co-watched by other users may also help to form user C's reference points for some implicit or explicit factors, such as the ``also-viewed'' or ``also-purchased'' links recommended by the platform. Along this way, we embody the advantage of graphical structures for reference-dependent preference learning, even for cold-start consumers with very limited historical consumption data (like user C).



While the reference network helps to address the challenge of reference points modeling, these reference points are identified at the product level rather than the product attribute level. As supported by the prior studies~\citep{wang2019incorporating}, however, the attribute level is a more appropriate granularity for reference points modeling as consumers tend to form different reference points when evaluating different attributes, and the proximity in reference network may not necessarily indicate proper reference points.
For instance, as indicated in Figure~\ref{fig:rdpnex}(b), the movie \emph{Dunkirk} is closely connected to \emph{Matrix} in the reference network; however, it would be inappropriate to take \emph{Dunkirk} as a valid source of reference point for user A, since she only focuses on Sci-Fi movies and \emph{Dunkirk} as a military movie can provide limited reference value in the genre-aspect. In this case, the product transitive relations provided by the reference network encounter problems; instead, we should drill down to the product attribute level, \emph{e.g.}, the genre of a movie, for accurately capturing the reference points for user A. Indeed, this argument is also supported by the Multiple Attribute Utility Theory (MAUT) where product evaluation should be decomposed into individual attribute levels, \emph{e.g.}, the genre, the director, or the actors of a movie, and followed by a compositional strategy to form the overall assessment~\citep{keeney1993decisions}.  Meanwhile, in quantitative marketing research, conjoint analysis approach is also used by assuming that consumers' preferences toward products can be aggregated from different product attributes separately in an additive way~\citep{kim2017benefit, kohli1990heuristics}. Inspired by these, we consider decomposing the reference network at the product attribute level for pursuing fine-grained reference points modeling. Specifically, we assume the formation of connecting edges could be ascribed to several salient aspects that characterize the products. As a result, we can decompose the edges according to the product attributes and generate the so-called \emph{Attributed Reference Network} (ARN) defined as follows.



\begin{definition}[Attributed Reference Network]\label{def:arn}
An attributed reference network $\mathcal{G}^k=<\mathcal{V}^k,\mathcal{E}^k>$ is a sub-network decomposed from the reference network $\mathcal{G}$ with respect to the $k$-th attribute, $k=1,\cdots, K$. 
The nodes in $\mathcal{G}^k$ represent all the purchased products, \emph{i.e.}, $\mathcal{V}^k=\mathcal{V}$, while only partial edges in $\mathcal{G}$ are preserved in $\mathcal{G}^k$ with $\mathcal{E}^k=\lbrace r_{i,j}  | r_{i,j} \in \mathcal{E} ~\&~ a_i^k \cong a_j^k \rbrace$, where $a_i^k$ and $a_j^k$ denote the attribute values of the nodes $v_i$ and $v_j$ \emph{w.r.t.} the $k$-th attribute, respectively, and $\cong$ indicates that $a_i^k$ and $a_j^k$ are in a same level (if real-valued) or exactly the same (if categorical). ~\hfill$\blacksquare$
\end{definition}

As depicted in Figure~\ref{fig:rdpnex}, the edge between two movies can be driven by either the \emph{genre} or the \emph{director} attribute, resulting in the two ARNs in Figure~\ref{fig:rdpnex}(c) and (d).  Through analyzing the structures of these ARNs, we can gain finer-granularity insights into how consumers could form diverse reference points at the attribute level in a data-driven manner. Suppose each product has $K$ important attributes, then a total of $K$ ARNs can be decomposed from the reference network with different connecting structures. It is worth noting that we can also relax the requirement of the symbol $\cong$ in Definition~\ref{def:arn} when needed to accommodate various scenarios.
For example, we can keep the edge between two movies in a genre-defined ARN as long as the two movies share similar categorical values, \emph{e.g.}, one belongs to both comedy and romance genres and the other is only a comedy. For another, 
when constructing the ARN \emph{w.r.t.} the ``actor'' attribute, we can preserve edges between movies sharing similar actors, \emph{e.g.}, action movie stars like \emph{Jackie Chan} and \emph{Arnold Schwarzenegger}. Consequently, most nodes in a particular ARN can be closely connected through higher-order relations, sharing the same or semantically similar attribute values.

By anchoring reference points in ARNs, reference-dependent preference can be measured as the utility gain by comparing a focal product and the reference points within the networks, which is called the \emph{interest-inspired preference} of consumers. On the other hand, product price, as a special attribute, also plays an indispensable role in shaping consumers' purchase decisions~\citep{ban2021personalized} but is often overlooked in prior work. In fact, the consumer preference can also be modeled by comparing against price-centric reference points, commonly referred to as the reference prices~\citep{lattin1989reference, winer1986reference}, to gauge the so-called \emph{price-inspired preference} of consumers in making purchase decisions. Building upon these motivations, we incorporate both interest-inspired and price-inspired preferences for modeling consumers' reference-dependent utility, which is formally defined as the problem of designing \emph{Reference-dependent Choice Model for Recommendation}  as follows.


\textsc{\textsf{Problem Definition~(Reference-dependent Choice Model for Recommendation)}}. Consider a merchant offering a diverse range of products represented by $\mathcal{V} = \lbrace v_{1}, \ldots, v_{|\mathcal{V}|} \rbrace $. Each product $v_i\in\mathcal{V}$ is characterized by $K$ different attributes denoted as $\mathbf{a}_{i} \in \mathbb{R}^{|K|}$ and is given a price $p_i$. The consumers who have ever purchased products from the merchant are represented by $\mathcal{U}=\left\{u_{1}, \ldots, u_{|\mathcal{U}|}\right\}$. The interactions between consumers and products are represented by $\mathcal{Y} = \{y_{u_i,v_j | u_i \in \mathcal{U}, v_j\in \mathcal{V}}\}$, where $y_{u_i,v_j}=1$ if $u_i$ has ever purchased $v_j$ and 0 others. Now given a consumer $u_i$ and a candidate product $v_j$ with unknown $y_{u_i,v_j}$, the task is to estimate $u_i$'s reference-dependent preference (utility) toward $v_j$ for personalized recommendation, where both the interest-inspired and price-inspired preferences should be retrieved from the ARNs at the attribute level and integrated for preference learning. 
~\hfill$\blacksquare$




\section{Reference-dependent Choice Model for Recommendation}

\begin{figure*}[t!]
	\centering
	\includegraphics[width=0.99\textwidth]{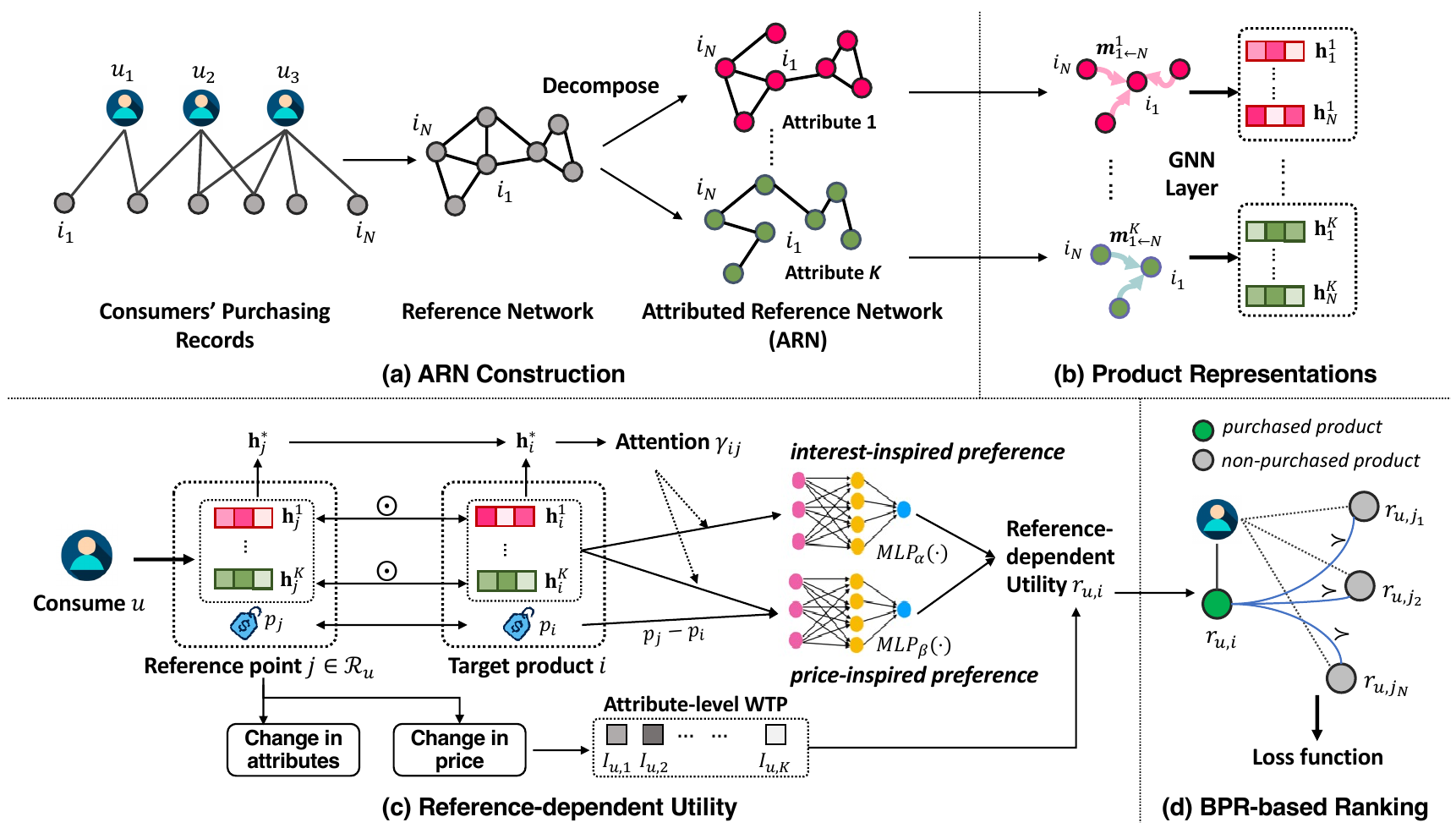}
	\caption{The architecture of \emph{ArcRec}.}
	\label{fig:framework}
\end{figure*}

In this section, we propose a reference-dependent choice model for personalized product recommendation. Our model addresses two pivotal issues that are pertinent to consumer choices, \emph{i.e.}, how to identify suitable reference points for each consumer regarding a target product, and how to construct consumers' reference-dependent utilities based on the reference points. In what follows, we first summarize the overall logic and the key modules of our model (see Figure~\ref{fig:framework}). 


As discussed in Section~\ref{sect:decomp}, the intuitive reference points for a consumer contain influences from both her historically purchased products and the crowd’s preferential information at the attribute level. To model the crowd’s preferential information, various ARNs are constructed upon the historical purchases of the crowds, and the product representations in a manifold space can then be derived through Graph Neural Networks (GNN), which together with a consumer's own purchase history can quantify the attribute-level reference points. Figure~\ref{fig:framework}(b) shows the module for product representation learning on ARNs, with the technical details presented in Section~\ref{sec:prodembed}.



We then calculate consumers' reference-dependent utilities by leveraging these attribute-level reference points and prices. Specifically, we decompose the utility into interest-inspired and price-inspired preferences, to account for the two major gauges in consumer choices, which can then be modeled by deep neural networks.
To assemble the attribute-level reference-dependent utilities into an integrated one, we design an Attribute-level Willingness-To-Pay (AWTP) measure in favor of product attributes that determine the price tolerance of the consumer towards the target product. Figure~\ref{fig:framework}(c) depicts the module for modeling a consumer's reference-dependent utility, with the technical details given in Section~\ref{sec:utility}. 


We finally integrate the reference-dependent utility computation into the Bayesian Personalized Ranking (BPR) framework for an end-to-end model training, as illustrated in Figure~\ref{fig:framework}(d) and Section~\ref{sec:train}. We name the whole model architecture in Figure~\ref{fig:framework} as \underline{A}ttributed \underline{R}eference-dependent \underline{C}hoice Model for \underline{Rec}ommendation (\emph{ArcRec}). The process of utilizing \emph{ArcRec} for product recommendation generation and its capability of addressing cold-start issues are introduced in Section~\ref{sec:recgen}.

\subsection{Product Representation Learning in Attributed Reference Network} 
\label{sec:prodembed}
ARNs are capable of capturing the various ways in which products are connected and referred based on different attributes. Traditional network representation methods such as DeepWalk~\citep{perozzi2014deepwalk}, node2vec~\citep{grover2016node2vec}, \emph{etc.}, can map nodes into a low-dimensional representation space. However, these methods rarely address higher-order relations between nodes and therefore hardly retain the transitivity of network structures, resulting in information loss. We here propose to use Graph Neural Network (GNN) \citep{welling2016semi, velivckovic2017graph}, a widely used deep learning framework, to preserve network structural properties effectively~\citep{wu2020comprehensive}. The underlying principle of GNN is to perform message passing and aggregation between adjacent nodes in the network, enabling the iterative and adaptive propagation of high-order information across the network. In this regard, we can treat reference relations between products as messages that propagate in the networks.

We first define node embeddings for each ARN. Specifically, we introduce an embedding layer to encode each product node $v_i\in\mathcal{V}$ in the $k$-th ARN $\mathcal{G}^k$ with a $d$-dimensional embedding vector $\mathbf{e}_i^k \in \mathbb{R}^d$, $d \ll |\mathcal{V}|$. Here, an embedding layer is a special MLP layer without bias and non-linear activation function, and $\mathcal{G}^k$ is denoted as an adjacency matrix $\mathbf{S}^k$, where $s_{i,j}^k=1$ if products $v_i$ and $v_j$ have been co-purchased and have a same level of attribute $k$ and 0 otherwise. It is worth noting that $\mathbf{S}^k$ is not necessarily a binary matrix but can be a real-valued one if we prefer weighted edges, \emph{e.g.}, using the frequency of a product pair's co-purchases. Given $\mathbf{S}^k$ for each attribute $k$, we can then perform message passing and aggregation to obtain node representations.

\emph{Message Passing.} 
According to the standard procedure of GNN-based models~\citep{welling2016semi, velivckovic2017graph}, messages will be propagated between connected neighbors to derive the node representations, which can generally be formulated as 
\begin{equation}
\label{eq:mp}
	\mathbf{m}_{i \gets j}^{k, (l+1)} =  \frac{1}{\sqrt{|\mathcal{N}_i^k|} \sqrt{|\mathcal{N}_j^k|}} \, \mathbf{h}_{j}^{k, (l)},
\end{equation}
where node $v_i$ is the focal node, node $v_j$ is one of its neighbors in $\mathcal{G}_k$ with $s_{i,j}^k=1$, and $l$ represents the $l$-th layer of the GNN. The term $\mathbf{m}_{i \gets j}^{k, (l+1)}$ is defined as the message passed from the neighbor node $v_j$ to the focal node $v_i$ following the existing path in $(l+1)$-th layer. The term $\mathbf{h}_{j}^{k, (l)}$ represents the latent embedding of node $v_j$ in $\mathcal{G}^k$ in the $l$-th layer. Note that when $l=0$, $\mathbf{h}_{j}^{k, (0)}$ denotes the initialized embedding of node $v_j$, \emph{i.e.}, $\mathbf{e}_i^k$. The normalization term $1 / (\sqrt{|\mathcal{N}_i^k|} \sqrt{|\mathcal{N}_j^k|})$ is the graph Laplacian norm to constrain the embedding scales, where $\mathcal{N}_i^k$ denotes the neighbor set of node $v_i$ in $\mathcal{G}^k$.

\emph{Message Aggregation.} With the messages passed between one-hop neighbors, we can aggregate all the messages to update the representation of the focal node. Apart from aggregating information from other nodes, we also add a residual connection \citep{vaswani2017attention} to preserve the node's own features. We name this message channel as \emph{self channel}, which copies node embeddings directly to the next layer. Taking node $v_i$ as an example, we have message vector $\mathbf{m}_{i \gets self}^{k, (l+1)} = \frac{1}{|\mathcal{N}_i^k|} \, \mathbf{h}_{i}^{k, (l)}$. The new representation of $v_i$ in the $(l+1)$-th layer can then be formulated as
\begin{equation}
\label{eq:ma}
	\mathbf{h}_{i}^{k, (l+1)} = \overbrace{\mathbf{m}_{i \gets self}^{k, (l+1)}}^{\text{self}} \ +  \overbrace{\sum_{j \in \mathcal{N}_i^k} \mathbf{m}_{i \gets j}^{k, (l+1)}}^{\text{from neighbors}}.
\end{equation}

\emph{Layer Combination.} 
Since the messages are propagated iteratively on GNN, each layer would capture information of different granularity. As a result, we further combine the node embeddings in different layers to generate the final representation with respect to the underlying network structure as follows:
\begin{equation}
\label{eq:lc}
	\mathbf{h}_{i}^{k} = \sum_{l=0}^{L} \frac{\theta_l}{\sum_{l=0}^{L} \theta_l}  \, \mathbf{h}_{i}^{k, (l)},
\end{equation}
where $\mathbf{h}_{i}^{k}$ is the encoded representation of product $v_i$ in the ARN $\mathcal{G}^k$, $L$ denotes the total number of layers, and $\theta_l$ represents the importance of the $l$-th layer embeddings. In this paper, we simply set $\theta_l=1/(l+1)$ instead of learning it through the attention network to reduce model complexity. 

By utilizing GNNs to encode each ARN, we obtain multiple embedding vectors for each product corresponding to different product attributes. These embedding vectors capture the attribute-level reference information from the past purchases of the crowds and are used as inputs to model consumers' reference-dependent choices in the following sections.

\subsection{Reference-dependent Choice Modeling via Deep Neural Network}
\label{sec:utility}

In this section, we leverage a consumer's purchase histories along with the obtained embedding vectors to model her reference-dependent choices. The underlying intuition is two-fold. On the one hand, the historically purchased products constitute part of her past experiences explicitly. On the other hand, the products' representations encoded by GNNs take into account the influence of ``wisdom of crowds'' implicitly. We thus combine these two key factors together to form reference points.

Specifically, given a consumer $u$ and a target product $i$\footnote{We here use $i$ to denote product $v_i$ for brevity.}, $\mathcal{R}_u$ represents the set of products that have been purchased by consumer $u$, $\mathbf{h}_i^k$ denotes the GNN encoded representation of product $i$ in the aspect of the $k$-th attribute, and $p_i$ denotes the price of product $i$. By retrieving the GNN encoded product representations in $\mathcal{R}_u$ as reference points and comparing these representations and prices with those of the target product, we formulate the basic utility function for consumer $u$ towards target product $i$ as
\begin{equation}\label{eq:demand_initial}
	r_{u,i} = \frac{1}{|\mathcal{R}_u|} \sum_{j \in \mathcal{R}_u} \sum_{k \in \mathcal{K}} f_{u,k}(\mathbf{h}_{i}^{k}, \mathbf{h}_{j}^{k}, p_i, p_j),
\end{equation}
where $f_{u,k}(\cdot)$ evaluates how consumer $u$ perceives the product utility by comparing the target product $i$ with one of the reference points from the aspect of attribute $k$. It is noteworthy that since each ARN shows diverse connection structures, $\mathbf{h}_{j}^{k}$ differs for each attribute, echoing existing empirical evidence that the selection of reference points may vary when evaluating different product attributes. Finally, the overall utility $r_{u,i}$ is obtained by taking the summation of the reference effects over different attributes in $\mathcal{K}$ for each reference point, and then taking the average of the reference effects on different reference points in $\mathcal{R}_u$. 


It is important to note that, while all the reference points contribute to the consumer utility computation in Equation~\eqref{eq:demand_initial}, their reference effects can be quite different and thus lead to the heterogeneity of reference points. In practice, consumers typically engage in an internal information search process when making purchasing decisions, relying on their memory of relevant information pertinent to the target product. During this process, some information is more easily activated and retrieved than others, suggesting that these reference points carry greater weights. Therefore, it is imperative to differentiate the significance of different reference points in Equation~\eqref{eq:demand_initial}. 

Motivated by the heterogeneity of reference points, we reformulate the utility function in Equation~\eqref{eq:demand_initial} by incorporating weights for reference points using an attention mechanism. This mechanism, originally introduced in machine translation tasks to identify salient positions of words in a sentence, encodes the most relevant information through an alignment model~\citep{bahdanau2014neural}. In our context of consumer choices, we use the attention mechanism to capture consumers' focal attention to different reference points, where reference points with higher attention weights are retrieved more easily and contribute more to the formation of reference effects for the target product. Formally, we parameterize $\gamma_{i,j}$ as the inner dot between product node embeddings as follows:
\begin{equation}
\begin{aligned}
\label{eq:product_att}
	\tilde{\gamma}_{i,j} &= {\mathbf{h}_{i}^{*}}^{\top} {\mathbf{h}_{j}^{*}}, \quad \mathbf{h}_{i}^{*} = \mathbf{h}_{i}^{1} \mathbin\Vert \cdots \mathbin\Vert \mathbf{h}_{i}^{K}, \\
	\gamma_{i,j} &= \text{Softmax}(\tilde{\gamma}_{i,j}) = \frac{\exp(\tilde{\gamma}_{i,j})}{\sum_{j^\prime \in \mathcal{R}_u} \exp(\tilde{\gamma}_{i,j^\prime})},
\end{aligned}
\end{equation}
where ``$\mathbin\Vert$'' is the concatenation operator, $\mathbf{h}_{i}^{*}$ combines the product node representations from $K$ ARNs and draws a holistic picture, and the softmax function is applied to normalize the attention scores. Now given the parameter $\gamma_{i,j}$, we can plug it as a weight for each pair of comparisons in Equation~\eqref{eq:demand_initial}, which gives the enhanced utility function as follows:
\begin{equation}\label{eq:demand_w1}
r_{u,i} = \sum_{j \in \mathcal{R}_u} \gamma_{i,j} \sum_{k \in \mathcal{K}} f_{u,k}(\mathbf{h}_{i}^{k}, \mathbf{h}_{j}^{k}, p_i, p_j).
\end{equation}
Hence, our remaining task is to specify the formulation of the attribute-level utility function $f_{u,k}(\cdot)$ to account for the reference point $j$-referenced utility of consumer $u$ to product $i$ in terms of the aspect of attribute $k$. 


\subsubsection{Reference-dependent Utility with Dual Preferences.}
As discussed previously in Section~\ref{sect:decomp}, reference-dependent utility could be modeled by comparing a target product against reference points in each salient attribute from two aspects, \emph{i.e.}, the \emph{interest-inspired preference} focused on the product properties, and the \emph{price-inspired preference} focused on product prices. We formulate these two aspects to implement $f_{u.k}(\cdot)$ in Equation~\eqref{eq:demand_w1} as follows. 

\textbf{Interest-inspired Preference.} We employ the \emph{Hadamard product} to compare a target product against a reference point under a given attribute $k$. Specifically, let $\mathbf{h}_i^k$ and $\mathbf{h}_j^k$ represent the target product $i$ and the reference point $j$, respectively, the Hadamard product is then formulated as $\mathbf{h}_{i}^{k} \odot \mathbf{h}_{j}^{k}$, in which a feature dimension could be intensified if the two product embeddings exhibit similar level of values (either positive or negative) in that specific dimension. 
Then, we apply a transformation with an MLP parameterized by $\mathbf{\alpha}$, \emph{i.e.}, $\text{MLP}_{\alpha} \left(\mathbf{h}_{i}^{k} \odot \mathbf{h}_{j}^{k}\right)$, to obtain consumers' personalized perceptions as the interest-inspired preference. This MLP transformation simulates the information processing process in consumers' comparison of a product against their own reference points. 

\textbf{Price-inspired Preference.} In real-life purchasing decisions, consumers also assess a target product by comparing its price with that of the reference points, which is commonly termed as \emph{reference price}. The influence of reference price on consumer choices has been extensively examined in the literature~\citep{lattin1989reference, winer1986reference}, typically through the integration of a ``sticker shock" term into the utility function. This term captures the disparity between the reference price and the shelf price of the evaluated product and is commonly assumed to have a linear additive effect. Formally, this term is formulated as $\beta \cdot (p_j - p_i)$, where $\beta$ represents consumers' price sensitivity, and $i$ and $j$ denote the target product and reference point, respectively. 

Prior research on reference price has predominantly concentrated on the product level~\citep{lattin1989reference, winer1986reference}. Nevertheless, consumers often evaluate the price of a product based on specific product attributes rather than considering them in aggregate~\citep{andrews1999mds}, implying that the price interacts extensively with other product attributes. For example, when purchasing a handbag, consumers may evaluate whether the brand is worth the price but overlook its functionality. Motivated by these observations, in what follows, we propose to integrate an attribute-aware ``sticker shock'' term into the reference-dependent utility function for price-inspired preference modeling. 


Specifically, we extend the parameter $\beta$ in $\beta \cdot (p_j - p_i)$ to the price sensitivity at the product attribute level. As evidenced by prior studies, consumers' attribute-level interest-inspired preferences will significantly affect their perceptions of price sensitivity~\citep{mellens1996review, ishak2013review}. For instance, \cite{krishnamurthi1991empirical} demonstrate that consumers with a strong interest for a specific brand are less price-sensitive than other consumers in making choice decisions. In this light, we assume that consumers have product attribute-aware price perceptions, and propose an \emph{attribute-aware price sensitivity} mechanism to capture this nuanced interactive relations. Specifically, we empower the parameter $\beta$ in $\beta \cdot (p_j - p_i)$ with another MLP transformation denoted as $\text{MLP}_{\beta} \left(\mathbf{h}_{i}^{k} \odot \mathbf{h}_{j}^{k}\right)$. This transformation aims to capture the influence of attribute-level interest-inspired preferences (represented by $\mathbf{h}_{i}^{k} \odot \mathbf{h}_{j}^{k}$ as discussed earlier) on a consumer's price perception. 

By integrating the models for both the interest- and price-inspired preferences into Equation~\eqref{eq:demand_w1}, the reference-dependent utility $r_{u,i}$ can be further formulated as
\begin{equation}
\label{eq:demand_att}
r_{u,i} = \sum_{j \in \mathcal{R}_u} \gamma_{i,j}  \sum_{k \in \mathcal{K}} \left(\overbrace{ \text{MLP}_{\alpha} \left(\mathbf{h}_{i}^{k} \odot \mathbf{h}_{j}^{k} \right)}^{\text{interest-inspired preference}} \right. 
\left. + \overbrace{\text{MLP}_{\beta} \left(\mathbf{h}_{i}^{k} \odot \mathbf{h}_{j}^{k}\right) \cdot \left(p_j - p_i\right)}^{\text{price-inspired preference}} \right).
\end{equation}

\subsubsection{Reference-dependent Utility with Attribute Salience.} 
 Equation~\eqref{eq:demand_att} implies all attributes are of equal importance in the evaluation of the target product, a condition that may not hold in real-world scenarios. For example, for our case in Figure~\ref{fig:rdpnex}, user A may be a Sci-Fi fan and thus is critical of the movie genre but indifferent to movie directors or actors. Indeed, the Multiple Attribute Utility Theory~\citep{keeney1993decisions} and the traditional conjoint analysis approach~\citep{jedidi2002augmenting} have highlighted the importance of considering possible trade-offs among product attributes in modeling consumers' choices. Motivated by these, we here extend this argument by positing that consumers' trade-offs among product attributes do exist and are personalized.

In the literature, while various recommendation methods have treated product attributes as side information to enrich product or consumer representations~\citep{rendle2010factorization, zhang2016collaborative, wang2018dkn}, few efforts have been devoted to modeling the personalized salience of product attributes. However, it is intuitive that the personalized choice of product attributes could be discovered from consumers' purchase records with both the interest and price considerations. For instance, fans of certain pop stars tend to prioritize the singer aspect and may rush to buy a newly released album from these stars despite a significant price premium. This implies that we can model a consumer's attribute choice by observing her attribute-level product purchases with price variations. With these motivations, we propose a novel measurement called \emph{Attribute-level Willingness To Pay} (AWTP) to capture the personalized salience of product attributes.

We consider the extent to which the product price will affect a consumer's demand for a particular product attribute as the measurement of the attribute importance. 
Specifically, given a consumer $u$, for each attribute $k$, we assume that the consumer has a base demand of attribute $k$ to fulfill her routine requirements, along with the corresponding price perception. The demand deviation can be defined as the deviation of a purchased product from the base demand on attribute $k$, and the price change can be defined analogously as the price deviation of a purchased product from the base price. In our approach, we implement the base as a center point with the average attribute level as well as the average price of the products previously purchased. Formally, the center point is defined as 
\begin{equation}
    \overline{\mathbf{h}_{u}^{k}}=\frac{\sum_{i \in \mathcal{R}_u} \mathbf{h}_{i}^{k}}{\left|\mathcal{R}_u\right|}, \quad \overline{p_u}=\frac{\sum_{i \in \mathcal{R}_u} p_i}{\left|\mathcal{R}_u\right|}.
\end{equation}

Then, we define the attribute demand deviation. Concretely, for each attribute $k$, we define the demand deviation as the deviation of each purchased product $i \in \mathcal{R}_u$ from the center point in the representation space, leveraging the node embeddings for the $k$-th attribute. This is achieved by computing the distance between the embedding $\mathbf{h}_i^k$ and the embedding of the center point $\overline{\mathbf{h}_{u}^{k}}$. To normalize the changes, we calculate the ratio of the deviation to the embedding of the center point, representing the relative change in attribute choices. Here, the Euclidean distance is used to measure the deviation of node embeddings in high dimensional space. Then, given a consumer's demand deviations in a specific attribute, we treat the price change ratios to the center point as an indicator for the willingness to pay for the attribute. Through aggregating the change ratios to the center point from all the past purchases, we can derive the AWTP score for consumer $u$ to attribute $k$ as follows:
\begin{equation}
\label{eq:attr_im}
\begin{aligned}
	&\tilde{I}_{u,k} = \sum_{i \in \mathcal{R}_u} \frac{\left|p_i-\overline{p_u}\right| \bigg/ \left|\overline{p_u}\right|}{\left|\left| \mathbf{h}_{i}^{k}-\overline{\mathbf{h}_{u}^{k}}\right|\right| \bigg/ \left|\left|\overline{\mathbf{h}_{u}^{k}}\right|\right|}, \\
	&I_{u,k} = \text{Softmax}(\tilde{I}_{u,k}) = \frac{\exp(\tilde{I}_{u,k})}{\sum_{k^\prime \in \mathcal{K}} \exp(\tilde{I}_{u,k^\prime})},
\end{aligned}
\end{equation}
where the Softmax function is utilized to normalize the AWTP values, facilitating the comparison across different attributes. A higher $I_{u,k}$ suggests the consumer $u$ may prioritize the product attribute $k$ since she is willing to accept a wider range of price fluctuation after controlling the changes from products themselves.


We finally integrate AWTP as a specially designed attention network into our utility function in Equation~\eqref{eq:demand_att}, and the ultimate reference-dependent utility is given by
\begin{equation}
\label{eq:demand_final}
r_{u,i} = \sum_{j \in \mathcal{R}_u} \gamma_{i,j}  \sum_{k \in \mathcal{K}} I_{u,k}  \left(\overbrace{ \text{MLP}_{\alpha} \left(\mathbf{h}_{i}^{k} \odot \mathbf{h}_{j}^{k} \right)}^{\text{Attribute-level interest preference}} \right. 
\left. + \overbrace{\text{MLP}_{\beta} \left(\mathbf{h}_{i}^{k} \odot \mathbf{h}_{j}^{k}\right) \cdot \left(p_j - p_i\right)}^{\text{Attribute-level price preference}} \right).
\end{equation}

\subsection{Ranking-based Recommendation and Model Optimization}
\label{sec:train}


In the field of recommender systems, it is widely acknowledged that providing consumers with a ranked list of recommendations is more appropriate than directly predicting product ratings~\citep{rendle2010pairwise, rendle2012bpr, he2018adversarial}. In response to this, a pairwise modeling framework called Bayesian Personalized Ranking (BPR) is introduced for generating product ranking lists for consumers~\citep{rendle2012bpr}. BPR offers a balanced approach between recommendation performance and computational complexity, and can be combined with traditional factorization models to infer consumer preferences from their implicit feedback. In our design, we integrate the constructed model for reference-dependent utility into the BPR framework to enable end-to-end learning of product representations and improve the overall performance of recommender systems.

A crucial assumption for BPR is that products with which consumers interact are considered to be more preferred than those they have not interacted with, which has been proven to be effective and has been widely adopted in many previous studies~\citep{ding2018improved, he2018adversarial}. In practice, the effectiveness of this approach is ensured by the randomization of training samples. To adhere to this strategy, we define the positive set and the negative set to construct the BPR-style loss as follows:
\begin{equation}
\label{eq:dataset}
	\mathcal{D}^+_u = \{i \,|\, i \in \mathcal{R}_u\}, \quad  \mathcal{D}_u^- = \{j \,|\, j \in \mathcal{V} \backslash \mathcal{R}_u\},
\end{equation}
where $\mathcal{R}_u$ is the product set that consumer $u$ has interacted with, $\mathcal{V}$ is the entire product set, and the math symbol ``$\backslash$'' is the operator for complementary set. Suppose there are two products $i$ and $j$, and consumer $u$ purchases product $i$ but misses product $j$. The partial relationship of consumer preference over the two products can be defined as $i \textgreater_{u} j$ in BPR, where ``$\textgreater_{u}$'' indicates the personalized ranking structure satisfying the following inequality:
\begin{equation} \label{eq:inequality}
	i \textgreater_{u} j \, \Rightarrow \, r_{u,i} \textgreater r_{u,j}.
\end{equation}

The overall objective function for representation learning can be formulated as maximizing the utility difference between the positive products sampled from $\mathcal{D}^+_u$ and the negative products sampled from $\mathcal{D}^-_u$, which is given by
\begin{equation}
\label{eq:loss}
 \mathcal{L} = \sum_{u \in \mathcal{U}} \sum_{\substack{i \in \mathcal{D}_u^+, \ j \in \mathcal{D}_u^-}} \log\sigma\left(r_{u,i} - r_{u,j}\right),
\end{equation}
where $\sigma(x)$ is a sigmoid transformation, \emph{i.e.}, $ \sigma(x)= \frac{1}{1+\exp(-x)}$.
Apart from the objective function defined in Equation \eqref{eq:loss}, we also apply another commonly adopted $L_2$ norm as parameter regularization to prevent overfitting. We adopt Adaptive Moment Estimation (Adam) \citep{kingma2014adam} to optimize the objective function and infer the parameters. In each iteration, we sample a mini-batch of consumer and product pairs with their corresponding $K$ adjacency matrices and product prices to update the model parameters. The training algorithm is detailed in Algorithm \ref{alg:Framwork}.

\begin{algorithm}[h]
    \scriptsize
	\caption{The training algorithm of \emph{ArcRec}.}
	\label{alg:Framwork}
	\begin{algorithmic}[1]
		\Require Purchasing records $\mathcal{R}$, consumer set $\mathcal{U}$, product set $\mathcal{V}$, as well as attribute set $\mathcal{A}$.
		\Ensure Neural network parameters. 
		\State Randomly initialize model parameters by normal distribution, construct $K$ ARNs $\{\mathcal{G}^1, \cdots, \mathcal{G}^k, \cdots, \mathcal{G}^K\}$ with adjacency matrices $\{\mathbf{S}^1, \cdots, \mathbf{S}^k, \cdots, \mathbf{S}^K\}$ based on $\mathcal{R}$ and $\mathcal{A}$.
		\While {not converged}
		     \State {Sample consumer $u$ from $\mathcal{U}$, positive product $i$ from $\mathcal{D}^+_u$, and negative product $j$ from $\mathcal{D}^-_u$.}
            \For{$\mathbf{S}^k \in \{\mathbf{S}^1, \cdots, \mathbf{S}^k, \cdots, \mathbf{S}^K\}$}
            \State Derive $\mathbf{h}_{i}^{k}$ according to Equation \eqref{eq:mp} - Equation \eqref{eq:lc}.
            \EndFor
	        \For{$k \in \{1, \cdots,k, \cdots, K\}$}
	            \State Obtain AWTP scores $I_{u,k}$ according to Equation \eqref{eq:attr_im}.
	        \EndFor
	        \State Initialize the recommendation scores of $(u,i)$ and $(u,j)$, $r_{u,i}=0, r_{u,j}=0$.
	        \For{$t \in \mathcal{R}_u \backslash \{i\} $}
	            \State Compute $\gamma_{i,t}$ and $\gamma_{j,t}$ according to Equation \eqref{eq:product_att}.
	        \EndFor
	        \State Obtain aggregated $r_{u,i}$ and $r_{u,j}$ based on $I_{u,k}$ and $\gamma$ respectively according to Equation \eqref{eq:demand_final}.
	        \State Compute the objective function $\mathcal{L}$ according to Equation \eqref{eq:loss} as well as $L_2$ norm. 
			\State Perform gradient descent and update model parameters through the optimization method Adam.
		\EndWhile \\
		\Return Learned model parameters.
	\end{algorithmic}
\end{algorithm}

\subsection{Product Recommendation Generation}
\label{sec:recgen}

We here explain how to apply our constructed reference-dependent choice model to make recommendations. Specifically, for a consumer $u$ and a target product $i$ that has been previously purchased by others, we retrieve the corresponding node embeddings $\mathbf{h}_i^k$ from each ARN $\mathcal{G}^k$ according to Equation~\eqref{eq:lc}. Subsequently, we utilize Equation~\eqref{eq:demand_final} to estimate the reference-dependent utility for the product. By computing the utility scores for all available candidate products, we can then rank the candidates accordingly, with those receiving the highest utility scores being recommended to the consumer.

A prevailing challenge faced by most traditional recommendation methods is their limited applicability to \emph{cold-start} products, \emph{i.e.}, products with scarce or even void purchasing records in history. Our proposed modeling framework \emph{ArcRec} indeed exhibits potentials in addressing this challenge due to its attribute-level expressiveness.
While the representations of cold-start products cannot be directly constructed from past purchases, we can establish connections between these products and existing ones in different ARNs via the observed attribute values. Specifically, we consider a cold-start product as a new node denoted by $c$, possessing an attribute vector $\mathbf{f}_c$. In the $k$-th ARN, we connect this node to existing nodes that share similar attribute values with $\mathbf{f}_c^k$, regardless of the co-purchasing records. With the model parameters learned in the training procedure, we can also obtain $c$'s node embedding $\mathbf{h}_c^k$ from the $k$-th ARN through a mean pooling operation among its one-hop neighbors. Formally, the above process can be described as
\begin{align}\label{eq:cold_score}
	\mathbf{h}_c^k = \frac{1}{|\mathcal{N}_c^k|} \sum_{i \in \mathcal{N}_c^k} \mathbf{h}_i^k, \, c \notin \mathcal{V},\ i \in \mathcal{V},
\end{align}
where $\mathcal{N}_c^k$ denotes the one-hop neighbor set of product node $c$ in the ARN $\mathcal{G}^k$. With the learned representation $\mathbf{h}_c^k$, we can compute the utility score $r_{u,c}$ through Equation \eqref{eq:demand_final}, which aligns with the utility computation of previously purchased products and fits the ranking-based recommendation.

\section{Empirical Evaluations}
\subsection{Evaluation Overview}
Evaluation is critical in validating the efficacy of new recommendation methods. Conducting controlled experiments by deploying the model in a real-world e-commerce platform is an ideal approach for evaluation. However, such experiments can be costly and require platform support. As a result, offline evaluation based on archival dataset or synthetic dataset is commonly adopted by researchers to evaluate the newly proposed recommendation approaches~\citep{he2017neural, li2017utility, he2019mobile}. The use of synthetic datasets offers the advantage of known true purchase probabilities and parameters of the data generating process, enabling the simulation of controlled environments. On the other hand, archival datasets provide a more realistic representation of real-world situations and data distributions. To demonstrate the robustness of our proposed method in modeling consumers' choices and providing recommendations, we evaluate our model under both synthetic and archival dataset settings.

In the evaluation, we first simulate the ground truth purchasing probabilities with respect to consumers' decision process, taking into account both product attributes and price. Then, we can perform simulation experiments to validate the modeling details of the proposed model and examine its efficacy in revealing consumers' decision patterns. The simulation experiment serves two primary purposes. On one hand, we can validate whether our model can accurately reconstruct the true purchasing probabilities, thereby validating its ability to capture consumers' choice behaviors; on the other hand, we can further reveal the effectiveness of the key modeling strategies, such as price-inspired preference modeling in understanding consumers' attitudes toward the prices of different products.

Meanwhile, we also perform experiments on a real-world e-commerce dataset. We can systematically validate the recommendation performances with a hold-out sampling strategy to split the data into training and testing set. Then, we can answer the following questions through extensive experiments. 1) Can the proposed method outperform the state-of-the-art methods in terms of recommendation accuracy? 2) Given its ability in handling cold-start products, does our method maintain superior model performances under this case? 3) What roles do different modeling modules of ArcRec play in improving the recommendation performance? 4) Can we gain insights from consumers' AWTP scores to better understand their fine-grained preferences? 5) What impact do different hyperparameters have on the performance of ArcRec?

\subsection{Baseline Models}
To demonstrate the superiority of the proposed recommendation method, we compare our proposed method \textbf{\textsf{ArcRec}} with the following \textbf{14} baselines based on implicit feedbacks. They cover a wide range of recommendation methods including classical matrix factorization (MF) based methods, deep learning-based and graph-based CF methods, as well as several methods taking product attributes and/or prices into consideration. These baseline methods are briefly introduced as follows. Note that we adopt the same baselines for both synthetic and real-world datasets.

\textsf{{NMF}} \citep{paatero1994positive} is a collaborative filtering approach that factorizes the consumer-product rating matrix into latent factors. \textsf{{BPR-MF}} \citep{rendle2012bpr} is a well-known recommendation method that addresses implicit feedback by optimizing the ranking loss function. \textsf{{NeuMF}} \citep{he2017neural} is a state-of-the-art neural matrix factorization model that incorporates multiple non-linear neural network layers to predict consumer-product ratings.
We also include \textsf{{DeepICF}} \citep{xue2019deep}, a neural item-based CF method that learns global item similarities using deep feedforward networks.

In the realm of graph-based collaborative filtering, \textsf{{NGCF}} \citep{wang2019neural} is one of the earliest methods that utilizes graph neural networks to capture high-order relations in the consumer-product bipartite graph. \textsf{{DGCF}} \citep{wang2020disentangled} focuses on modeling relations between consumers and products by leveraging neighbor routing and disentangled graph neural network learning. \textsf{{LightGCN}} \citep{he2020lightgcn} extends NGCF by removing non-linear activation functions and transformation matrices, resulting in a linear GNN-based recommendation method.

In addition, we include recommendation methods that consider product attributes and price as benchmarks for our proposed methods. Specifically, \textsf{{WDL}} \citep{cheng2016wide}, \textsf{{DeepFM}} \citep{guo2018deepfm}, \textsf{{xDeepFM}} \citep{lian2018xdeepfm}, and \textsf{{AutoInt}} \citep{song2019autoint} treat product price as a continuous value directly, which aligns with our approach. However, these methods view price as a simple product attribute and focus on modeling feature interactions.

\begin{table}[t!]
	\centering
	\caption{Overview of baseline methods}
    \resizebox{0.99\textwidth}{!}{
		\begin{tabular}{c|ccccc}
			\toprule
			\textbf{Methods} & \textbf{Matrix Factorization} & \textbf{Deep Learning} & \textbf{GNNs} & \textbf{Attribute} & \textbf{Price} \\
			\hline
			NMF, BPR-MF & $\surd$ & & & & \\
			NeuMF & $\surd$ & $\surd$ & & & \\
			DeepICF & $\surd$ & $\surd$ & & & \\
			\hline
			NGCF, DGCF, LightGCN &  & $\surd$ & $\surd$ & & \\
			\hline
			WDL, DeepFM, xDeepFM, AutoInt & & $\surd$ & & $\surd$ & $\surd$ \\ 
			C-FMF & $\surd$ & & & $\surd$ & $\surd$\\
			PaDQ & $\surd$ & & & $\surd$ & $\surd$\\
			PUP & & $\surd$ & $\surd$ & $\surd$ & $\surd$\\ 
			\bottomrule
		\end{tabular}
	}
	\label{tab:baselines}
\end{table}

It is worth noting that several methods have been developed to exclusively incorporate product price into recommender systems. \textsf{{C-FMF}} \citep{wan2017modeling} is based on a feature-based matrix factorization framework with a focus on price, considering price as continuous values. \textsf{{PaDQ}} \citep{chen2014does} and \textsf{{PUP}} \citep{zheng2020price} also incorporate price into their recommendation methods, but they require splitting the price into different bins. However, determining the optimal number of bins and the best splitting method can be challenging and lacks customization for different datasets. In particular, \textsf{{PUP}} constructs a heterogeneous network by incorporating consumer, product and price as different types of nodes respectively. Then, a graph neural network is applied to learn node representations and make final recommendations. The above baseline methods are summarized in Table \ref{tab:baselines}. 

\subsection{Evaluation with Simulations}

\subsubsection{Simulation Setup.}
We draw on theoretical groundings in marketing science to design the data generation process. Specifically, we simulate a retailer with an assortment of $|\mathcal{V}|$ products and $|\mathcal{U}|$ consumers, where $|\mathcal{V}|$ and $|\mathcal{U}|$ are drawn from a discrete uniform distribution $X \sim \text{U}(1000, 2000)$. Each product $i$ is characterized by three categorical attributes $\{a_1, a_2, a_3\}$ and the product price. For each categorical attribute $a_n$, the value $i_{a_n}$ is drawn from a discrete distribution where the number of choices follows a discrete uniform distribution $X \sim \text{U}(3, 15)$, which is similar to rolling dice. The price $p_i$ is drawn from a standard normal distribution $X \sim \text{N}(0, 1)$ and transformed using an exponential function $\exp(p_i)$ to ensure non-negative prices. Correspondingly, each consumer $u$ is described by a few attribute preference parameters as well as a price sensitivity parameter \citep{andrews1998simulation, kim1995modeling}. For each attribute preference parameter $\alpha_{u, i_{a_n}}$,
we follow prior research \citep{andrews1998simulation} to draw it from a standard normal distribution $X \sim \text{N}(0, 1)$. 
The price sensitivity parameter $\beta_u$ is assumed to be log-normally distributed \citep{kim1995modeling}, such that $\beta_u = -\exp(\gamma)$, where $\gamma$ follows a standard normal distribution $X \sim \text{N}(0, 1)$. Then, the utility of consumer $u$ for product $i$ is given by Equation \eqref{eq:simulation}. 
\begin{equation}
\label{eq:simulation}
    r_{u,i} = \sum_{a_n} \alpha_{u, i_{a_n}} \cdot i_{a_n} + \beta_u \cdot (p_{u,r} - p_i) + \epsilon_{u,i},
\end{equation}
where $p_{u,r}$ refers to the consumer's reference price and $\epsilon_{u,i}$ is a random noise drawn from $X \sim \text{N}(0, 0.05)$.

The additional term $ \beta_u \cdot (p_{u,r} - p_i)$ is the ``sticker shock'' term proposed in \citep{winer1986reference} to model the impact of the reference price. Initially, when the consumer has not made any purchases, the product set $\mathcal{R}_u$ interacted by consumer $u$ is empty. In this case, we set $p_{u,r}$ to be the average price of all products in the retailer's assortment, calculated as $\frac{1}{|\mathcal{V}|}\sum_{j \in \mathcal{V}} p_j$, following the strategy used in \citep{zhang2019agent}. 
As the consumer learns and updates their internal knowledge through the purchasing process, we update the reference price $p_{u,r}$ as the average price of the prior purchases, computed as $p_{u,r}=\frac{1}{|\mathcal{R}u|} \sum_{j \in \mathcal{R}_u} p_j$, where $\mathcal{R}_u$ represents the set of prior purchases made by consumer $u$.
Since product choice often follows a multinomial logit model \citep{guadagni1983logit}, based on the defined utility $r_{u,i}$, the probability that consumer $u$ would choose product $i$ becomes:
\begin{equation}
\label{eq:simu_logit}
    P_{u,i} = \frac{\exp(r_{u,i})}{\sum_{j \in \mathcal{V}} \exp(r_{u,j})}.
\end{equation}

In each simulation period, consumers can make purchase decisions based on the generated purchasing probability $P_{u,i}$, and the total active periods for each consumer follow a discrete uniform distribution $X \sim \text{U}(5, 25)$. These simulation periods are used for model training and an additional period is also simulated for model evaluation, i.e., evaluating the recommendation performances.

\subsubsection{Evaluation Procedures.}
To evaluate the performance of the proposed model on a holdout test data, we designate the products from the last additional simulation period as the test data, while the remaining interactions are used as the training and validation data for optimizing our model as well as the baseline models. Since the parameters of the data generation process are known, we can obtain the true ranking list of products for each consumer in the test data according to the ground truth purchasing probabilities. 
Considering that the recommendation task is essentially a ranking problem, we compare the ranking results with the ground truth ranking list to evaluate model performance.
Two widely used metrics for evaluating the correlations between rank variables are adopted named Kendall rank correlation coefficient (Kendall's $\tau$) and Spearman's rank correlation coefficient (Spearman's $\rho$), which provide valuable insights into the degree of agreement between the predicted rankings and the ground truth rankings. Given the ground truth ranking list $\mathbf{T}$ and the recommended ranking list $\mathbf{C}$, The Kendall's $\tau$ and Spearman's $\rho$ are computed according to Equations~\eqref{eq:kendall} and \eqref{eq:Spearman} respectively, where $P$ is the number of concordant pairs between $\mathbf{T}$ and $\mathbf{C}$, and $d_i$ denotes the rank difference for product $i$ between $\mathbf{T}$ and $\mathbf{C}$. 
\begin{equation}
\label{eq:kendall}
    \text{Kendall's} \ \tau = \frac{1}{{|\mathcal{U}|}} \sum_{u \in \mathcal{U}} \left( \frac{4P}{|\mathcal{V}||\mathcal{V}-1|} -1\right),
\end{equation}
\begin{equation}
\label{eq:Spearman}
    \text{Spearman's} \ \rho = \frac{1}{{|\mathcal{U}|}} \sum_{u \in \mathcal{U}} \left( 1- \frac{6 \sum_{i \in \mathcal{V}} d_i^2}{|\mathcal{V}|^3 - |\mathcal{V}|} \right).
\end{equation}

Specifically, for a fair comparison, the embedding dimension $d$ is set to be 64 for all the models. In calibrating other hyperparameters of the proposed method, we set the mini-batch size, the learning rate of the optimization method Adam to be 500 and 0.003. To avoid overfitting, we apply the $L_2$ norm in the loss function and the coefficient is set to be 0.0001. Furthermore, the number of GNN layers $L$ is set to be 2 according to the grid search. Likewise, as for the benchmark methods, the hyperparameters are carefully tuned according to their best performances on the validation set.

\subsubsection{Product Ranking Performances.}
\begin{figure}[t]
	\centering
	\includegraphics[width=0.995\textwidth]{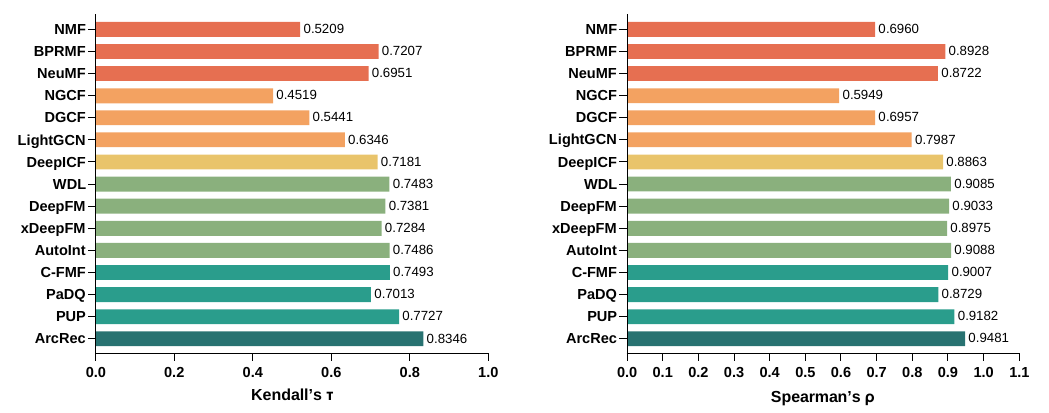}
	\caption{Ranking performances on synthetic data.}
	\label{fig:ranking_simu}
\end{figure}

We first conduct comprehensive comparison with the state-of-the-art methods in terms of product ranking based on the estimated utility, which is measured by the overall ranking quality via Kendall's $\tau$ and Spearman's $\rho$. The results are displayed in Figure \ref{fig:ranking_simu}, with methods in different categories plotted using different colors. Overall speaking, our method ArcRec performs consistently better than all the baselines in terms of both metrics. Compared with those traditional collaborative filtering methods such as NeuMF and LightGCN, our method achieves significant improvements, demonstrating the advantages of introducing attribute-level reference points as well as reference-dependent preference modeling. Compared with those baselines considering prices and/or attributes such as DeepFM and PUP, our method also shows superior performances in revealing consumers' inherent utility for the products, indicating that our method is a more appropriate and efficient way to exploit attribute and price information to capture consumers' inherent preferences. We will proceed to conduct more empirical evaluations to verify this point in the next section. 

\subsubsection{Consumers' Price Sensitivity Analysis.}
In the simulation of consumers' purchasing decision, we assign each consumer $u$ a coefficient ${\beta}_u$ to describe her personalized price sensitivity in Equation \eqref{eq:simulation} following existing research \citep{andrews1998simulation, kim1995modeling}. While our model proposes to capture consumers' attribute-aware price sensitivity, it is highly correlated with ${\beta}_u$ and can also reflect consumers' overall price sensitivity to some extent.
Thus, in this section, we further validate whether the consumers' heterogeneity over price can be captured by our method through a controlled experiment.

In our controlled experiment, we aim to evaluate the sensitivity of different recommendation methods to price changes among two groups of consumers with distinct levels of price sensitivity ${\beta}_u$. We randomly select 50 consumers with lower price sensitivity as a group $\boldsymbol{g}_{s}$ and 50 consumers with higher price sensitivity as another group $\boldsymbol{g}_{l}$. To examine the impact of price changes on recommendation rankings, we create a candidate product set $\mathcal{C}$ consisting of 30 randomly sampled products. 

Specifically, in our experiment, we consider the price change of a product as a treatment. For each consumer-product pair $(u,i)$, we examine the effect of the price change on the ranking of product $i$ in the consumer's ranking list. We define the treatment effect, denoted as $\textsf{TE}_{u,i}$, as the difference in the ranking of product $i$ before and after the treatment. The control price, denoted as $p_c$, represents the original price of product $i$, and the treatment price, denoted as $p_t$, represents the changed price.
Formally, the treatment effect for $(u,i)$ can be calculated using Equation \eqref{eq:TE_ui}, where $\textsf{rank}_{u,i}^{p_t}$ represents the ranking of product $i$ in the consumer's ranking list under the treatment price $p_t$. 
\begin{equation}
\label{eq:TE_ui}
    \textsf{TE}_{u,i}=\textsf{rank}_{u,i}^{p_t} - \textsf{rank}_{u,i}^{p_c},
\end{equation}
We aim to estimate the average treatment effect, denoted as $\textsf{ATE}$, for multiple $(u,i)$ pairs within a specific consumer group $\boldsymbol{g}_{s}$ or $\boldsymbol{g}_{l}$. The average treatment effect for group $\boldsymbol{g}_s$ is calculated using Equation \eqref{eq:TE}, denoted as $\textsf{ATE}_{s}$. The $\textsf{ATE}_l$ for the more price sensitive group $\boldsymbol{g}_{l}$ can be computed in a similar way.
\begin{equation}
\label{eq:TE}
    \textsf{ATE}_{s}=\frac{1}{|\boldsymbol{g}_s|} \sum_{u \in \boldsymbol{g}_s} \left( \frac{1}{|\mathcal{C}|} \sum_{i \in \mathcal{C}} \textsf{TE}_{u,i} \right).
\end{equation}

For an ideal recommendation method, we expect the average treatment effects for the two groups of consumers to exhibit certain patterns since the ranking results should be altered to respond to consumers' price concerns. When the product price decreases, the treatment effects should be negative, indicating that treated products with lower price may be ranked higher. Conversely, when the product price increases, the treatment effects should be positive, indicating a decline in the treated product's ranking. Additionally, we anticipate that the absolute value of $\textsf{ATE}_{s}$ will be smaller than that of $\textsf{ATE}_{l}$, as consumers in the $\boldsymbol{g}_s$ group have lower price sensitivity.

We conduct controlled experiments by repeating the process 30 times with randomly selected consumer groups and candidate product set $\mathcal{C}$. The ATE and their standard deviations are calculated for different recommendation methods, and the results are presented in Table \ref{tab:price_se}. We consider two types of treatments: increasing the product price by 10\% and decreasing the product price by 10\%, denoted as ``+10\%" and ``-10\%", respectively. Note that we have not included the methods PaDQ and PUP in the comparisons because they regard product price as discrete bins, making them not applicable for numerical manipulation in this particular experiment. 

From the results in Table \ref{tab:price_se}, we can draw two main conclusions. First, most of the methods estimate negative treatment effects when the product price is decreased and positive treatment effects when the product price is increased, aligning with our expectations for an ideal recommendation method. Second, our method achieves larger treatment effects (in absolute value) for the more price-sensitive group $\boldsymbol{g}_l$ and smaller treatment effects for group $\boldsymbol{g}_s$, demonstrating its ability to accurately distinguish consumers with different price sensitivities. In contrast, the baseline methods WDL, DeepFM, xDeepFM, and AutoInt estimate similar treatment effects for different consumer groups, while C-FMF even provides estimates that contradicted the ideal case. The findings highlight the effectiveness of our proposed method in accurately capturing consumers' underlying price sensitivities.

\begin{table}
	\renewcommand\arraystretch{1.1}
	\centering
	\caption{Treatment effects for consumers with different price sensitivity.}
	\label{tab:price_se}
		\begin{tabular}{@{}l|cc|cc@{}}
			\toprule
			\multirow{2}{*}{Method} & \multicolumn{2}{c|}{{\it -10\%}} & \multicolumn{2}{c}{{\it +10\%}} \\
			\cline{2-5}
			& $\beta_{small}$ & $\beta_{large}$ & $\beta_{small}$ & $\beta_{large}$ \\ \hline
			WDL & -5.68 ($\pm$ 0.01) & -5.71 ($\pm$ 0.00) & 5.75 ($\pm$ 0.01) & 5.78 ($\pm$ 0.00)  \\
			DeepFM & -6.27 ($\pm$ 0.01) & -6.22 ($\pm$ 0.01) & 5.12 ($\pm$ 0.01) & 5.25 ($\pm$ 0.01) \\
			xDeepFM & 0.76 ($\pm$ 0.01) & 0.75 ($\pm$ 0.01) & -0.75 ($\pm$ 0.01) & -0.76 ($\pm$ 0.01) \\
			AutoInt & -5.15 ($\pm$ 0.01) & -5.18 ($\pm$ 0.01) & 5.28 ($\pm$ 0.01) & 5.31 ($\pm$ 0.00) \\
			C-FMF & -22.55 ($\pm$ 0.92) & -14.28 ($\pm$ 0.14) & 22.45 ($\pm$ 0.42) & 14.23 ($\pm$ 0.17) \\
			\cline{1-5}
			ArcRec & \textbf{-14.50 ($\pm$ 1.06)} & \textbf{-20.51 ($\pm$ 0.45)} & \textbf{14.10 ($\pm$ 1.02)} & \textbf{19.42 ($\pm$ 0.44)} \\ \bottomrule
	\end{tabular}
\end{table}

\subsection{Empirical Evaluations on Real-world Data}
\label{sec:real}
\subsubsection{Data Description.}
We further validate the performances of our model using a real-world dataset named \textit{Cosmetics}, which is collected from an online cosmetic store in Taobao, one of the largest e-commerce platforms in China. The dataset comprises two data sources: consumer transaction data and basic product information. The transaction data contains the products a consumer bought in each transaction. The product information includes product attributes such as the retailing price, brand, category and size. These products are mainly skin care and make-up products including facial cleansers, facial masks, lotion, perfume, etc. We use brand, category and size as attributes to construct the ARNs. The detailed statistics of \textit{Cosmetics} are shown in Table \ref{tab:real_data}.

\begin{table}[t!]
	\renewcommand\arraystretch{1.05}
	\centering
	\caption{Description of Cosmetics dataset.}
	\footnotesize
	\resizebox{0.98\linewidth}{!}{
		\begin{tabular}{c|c|c|c|c|c}
			\toprule
			Dataset & \# of Consumers & \# of Products & \# Interactions & Attributes & \# Types of Attributes \\
			\midrule
			Cosmetics & 17,590 & 6,989 & 204,407 & Brand, category, size, price & 3 \\  
			\bottomrule
		\end{tabular}}
	\label{tab:real_data}
\end{table}

\subsubsection{Experimental Setup.}
An important difference from the simulation study is that the ground truth utility remains unknown in the real-world scenarios. Hence, both the proposed method ArcRec and the baseline methods are evaluated by following the standard evaluation protocol for recommendation with implicit feedbacks, i.e., leave-last-one-out strategy \citep{he2017neural, kang2018self, liu2021beyond}. To be specific, we hold-out the last purchased product of a consumer and the other remaining purchased products are used for training and validation. Then, the left one product is mixed with all the other non-purchased products to form the test set. The model evaluates the utility score for each product in the test set and ranks all products according to the estimated scores.

Given the rank of the candidate products according to the estimated scores, we firstly adopt the commonly used measurement hit ratio (HR) to evaluate how likely the recommended products would be actually purchased by the consumers. Formally, for each consumer $u$, a product $i$ is masked as the testing sample, and the top-$K$ product list $\mathbf{N}_u^K$ is recommended. The hit ratio can be measured by an indicator function, $g_{u} = \mathbb{I}(i \in \mathbf{N}_u^K)$, and the metric $\text{HR}@K$ can be computed in Equation \eqref{eq:hr}. Note that since we only have one single testing product for each consumer, $\text{HR}@K$ is equivalent to $\text{Recall}@K$ and is also proportional to $\text{Precision}@K$, so we only present the hit ratio results.
\begin{equation}
\label{eq:hr}
    \text{HR}@K = \frac{\sum_{u \in \mathcal{U}} \mathbb{I}(i \in \mathbf{N}_u^K)}{|\mathcal{U}|}.
\end{equation}

Except for the hit ratio, normalized Discounted Cumulative Gain (nDCG) is another widely used measurement to evaluate the recommendation performance. HR measures whether the ground truth product is present in the recommendation list, while nDCG accounts for the positions of hits. Specifically, given the top-$K$ ranked products for consumer $u$, the normalized Discounted Cumulative Gain (nDCG) is computed as follows,
\begin{equation}
\label{eq:ndcg}
\begin{aligned}
    & \text{DCG}@K(u) =  \sum_{p=1}^K \frac{2^{rel_p}-1}{\log_2(p+1)}, \\  
    & \text{nDCG}@K(u) = \frac{\text{DCG}@K(u)}{\text{IDCG}@K(u)}, \\
    & \text{nDCG}@K =  \frac{\sum_{u \in \mathcal{U}}\text{nDCG}@K(u)}{|\mathcal{U}|},
\end{aligned}
\end{equation}
where $rel_p=1$ if the $p$-th product in the recommendation list is actually purchased by the consumer $u$, and $rel_p=0$ otherwise. $\text{IDCG}@K(u)$ is the ideal $\text{DCG}@K(u)$, \textit{i.e.}, the maximum possible values of $\text{DCG}@K(u)$ with ideal rank of the recommendation list. As for hyperparameters tuning, we employ the same strategy as used in the simulation.

\subsubsection{Performance Comparison for Product Recommendation.} \label{sec:q3}
\begin{table}[t]
	\centering
	\caption{Product recommendation performances on real-world data.}
	\label{tab:rec_real}
	\begin{threeparttable}
		\begin{tabular}{@{}l|ccc|ccc@{}}
			\toprule
			Method & HR@5 & HR@10 & HR@15 & nDCG@5 & nDCG@10 & nDCG@15 \\  \midrule
			NMF & 0.0799 & 0.1209 & 0.1533 & 0.0520 & 0.0652 & 0.0738 \\
			BPR-MF & 0.0744 & 0.1234 & 0.1578 & 0.0469 & 0.0625 & 0.0716 \\
			NeuMF & 0.0839 & 0.1315 & 0.1702 & 0.0545 & 0.0697 & 0.0799   \\
			\cmidrule{1-7}
			DeepICF & 0.0748 & 0.1178 & 0.1459 & 0.0492 & 0.0624 & 0.0705 \\ 
			\cmidrule{1-7}
			NGCF & 0.0792 & 0.1242 & 0.1591 & 0.0520 & 0.0665 & 0.0757  \\
			DGCF & 0.0833 & 0.1333 & 0.1710 & 0.0531 & 0.0692 & 0.0791 \\ 
			LightGCN & \underline{0.0862} & 0.1344 & 0.1703 & \underline{0.0560} & \underline{0.0717} & 0.0811 \\ 
			\cmidrule{1-7}
			WDL & 0.0664 & 0.1038 & 0.1374 & 0.0433 & 0.0553 & 0.0642 \\ 
			DeepFM & 0.0655 & 0.1003 & 0.1320 & 0.0435 & 0.0546 & 0.0630 \\ 
			xDeepFM & 0.0655 & 0.1055 & 0.1400 & 0.0422 & 0.0550 & 0.0641 \\ 
			AutoInt & 0.0677 & 0.1073 & 0.1428 & 0.0454 & 0.0582 & 0.0675 \\ 
			C-FMF & 0.0708 & 0.1142 & 0.1513 & 0.0459 & 0.0598 & 0.0696 \\ 
			PaDQ & 0.0811 & 0.1237 & 0.1565 & 0.0530 & 0.0667 & 0.0754 \\ 
			PUP & 0.0851 & \underline{0.1350} & \underline{0.1744} & 0.0550 & 0.0709 & \underline{0.0814} \\ 
			\cmidrule{1-7}
			ArcRec & \textbf{0.0943} & \textbf{0.1442} & \textbf{0.1836} & \textbf{0.0614} & \textbf{0.0773} & \textbf{0.0875}  \\
			Improvement & 9.40\% & 6.81\% & 5.28\% & 9.64\% & 7.81\% & 7.49\% \\ \bottomrule
	\end{tabular}
	\begin{tablenotes}
      \item[1] The products to be recommended are ranked among the entire set of products without negative sampling.
    \end{tablenotes}
	\end{threeparttable}
\end{table}

We first perform a comparative evaluation to validate the superiority of our model over the benchmark methods in terms of product recommendation accuracy. The evaluation metrics used are $\text{HR}@K$ and $\text{nDCG}@K$, where the number of recommendation size $K$ is varied from 5 to 15 to demonstrate the robustness of the method. The comparative results are presented in Table \ref{tab:rec_real}, from which we can clearly find that our proposed method ArcRec consistently outperforms all the baselines. Upon closer examination of the model performances, the following observations can also be made.

Upon comparison of NeuMF with its counterparts BPR-MF and NMF, we observe that incorporating deep neural networks can significantly enhance recommendation performance, which is consistent with existing literature \citep{he2017neural, liu2021beyond}. Additionally, the graph-based CF methods can be viewed as extensions of classical matrix factorization-based methods, leveraging graph neural networks. In general, these graph-based CF methods (DGCF, LightGCN, etc.) demonstrate substantial improvements over the base model BPR-MF, highlighting the advantages and importance of modeling graph structures in recommender systems. However, these graph-based CF methods focus on constructing representations solely based on the consumer-product bipartite graph, without considering rich product attributes and reference dependent product relations. As a result, their recommendation performance falls short compared to our proposed method. 

Furthermore, it can be observed that the two methods PaQD and PUP incorporating price as a crucial module generally perform better than the simplified variants such as MF and NGCF, which demonstrates the effectiveness of price in modeling consumers' decision-making process. However, both methods require price discretization, which may lead to information loss since close product prices may be assigned to different bins, potentially overlooking the nuances of price differences. Conversely, our method treats product price as continuous value directly, eliminating the need for manual feature engineering and avoiding potential information loss. Interestingly, most recommendation methods that consider attributes and price perform even worse than pure collaborative filtering methods like NeuMF and DeepICF. This suggests that despite incorporating related information, these methods still struggle to effectively capture consumers' attribute-level interest- and price-inspired preferences in real-world datasets. In contrast, our proposed method offers a novel perspective for attribute modeling through the constructed reference network structure and reference-dependent choice model, enabling better utilization of side information. For instance, ArcRec outperforms xDeepFM by approximately 44\% (in terms of $\text{HR}@5$) and 45\% (in terms of $\text{nDCG}@5$), reinforcing the conclusion drawn from the simulation study.

\subsubsection{Performance for Recommending Cold Start Products.} \label{sec:q3}
To evaluate the performance of our method in recommending cold-start products, we conduct an experiment using a standard offline evaluation procedure for cold-start products~\citep{schein2002methods, barjasteh2016cold}. Due to the lack of valid online feedback on real cold-start products, we simulate this scenario by randomly holding out 1.5\% of the products in the dataset and treating them as new products with no purchasing records. These cold-start products are also excluded from the training stage.
In the testing phase, these cold-start products are connected to the nodes in ARNs based on their attribute values. According to Equation \eqref{eq:cold_score} and Equation \eqref{eq:demand_final}, we obtain utility scores for each consumer toward each cold-start product. Subsequently, we rank all the new products based on their utility scores, and generate a top-$K$ product list $\mathbf{N}_u^K$ for each consumer. We use $\text{HR}@K$ and $\text{nDCG}@K$ as evaluation metrics to measure the performance of the model in recommending cold-start products.

It is worth mentioning that several benchmark methods are not applicable to the cold start setting because new products do not appear in the interaction records, making it impossible to directly estimate the utility scores. Among all the baselines summarized in Table \ref{tab:baselines}, only five methods named WDL, DeepFM, xDeepFM, AutoInt and C-FMF can deal with this situation.
As shown in Table \ref{tab:cold_start}, it is evident that our proposed method ArcRec consistently outperforms other attribute and price-based recommendation methods. Among the baselines, C-FMF performs the second best when $K$ is small. However, it tends to be the worst when $K$ becomes larger. On the other hand, xDeepFM shows competitive results when $K$ is large. The results highlight the superiority of the proposed method in addressing cold start product recommendations, indicating its ability to capture consumers' utility at the attribute-level, as only attribute information is available in the absence of historical interaction data.

\begin{table}
	\renewcommand\arraystretch{1.1}
	\centering
	\caption{Recommendation performances for cold start products in the real-world data.}
	\label{tab:cold_start}
    \resizebox{0.99\textwidth}{!}{
	\begin{threeparttable}
		\begin{tabular}{@{}l|cccc|cccc@{}}
			\toprule
			Method & HR@5 & HR@10 & HR@15 & HR@20 & nDCG@5 & nDCG@10 & nDCG@15 & nDCG@20 \\ 
			\midrule
			WDL & 0.2321 & 0.3778 & 0.5099 & 0.5827 & 0.1539 & 0.1982 & 0.2280 & 0.2466 \\ 
			DeepFM & 0.2348 & 0.3875 & 0.4890 & 0.5756 & 0.1537 & 0.2004 & 0.2255 & 0.2446 \\
			xDeepFM & 0.2418 & \underline{0.4008} & \underline{0.5172} & \underline{0.6006} & 0.1565 & 0.2045 & \underline{0.2321} & \underline{0.2503} \\ 
			AutoInt & 0.2475 & 0.3965 & 0.4973 & 0.5861 & 0.1589 & \underline{0.2085} & 0.2295 & 0.2496 \\
			C-FMF & \underline{0.2568} & 0.3541 & 0.4305 & 0.4973 & \underline{0.1700} & 0.1981 & 0.2185 & 0.2329 \\ 
			\midrule
			ArcRec & \textbf{0.2983} & \textbf{0.4799} & \textbf{0.5807} & \textbf{0.6500} & \textbf{0.1759} & \textbf{0.2275} & \textbf{0.2529} & \textbf{0.2686} \\
			Improvement & 16.16\% & 19.74\% & 12.28\% & 8.23\% & 3.47\% & 9.11\% & 8.96\% & 7.31\% \\  \bottomrule
	\end{tabular}
	\begin{tablenotes}
      \item [1] The cold start products to be recommended are ranked among all the cold start products rather than the entire product set.
    \end{tablenotes}
    \end{threeparttable}}
\end{table}

\subsubsection{Ablation Study.} \label{sec:q3}

\begin{table}
	\centering
	\caption{Results for ablation studies on real-world data.}
	\label{tab:ablation}
		\begin{tabular}{@{}l|ccc|ccc@{}}
			\toprule
			Method & HR@5 & HR@10 & HR@15 & nDCG@5 & nDCG@10 & nDCG@15 \\ 
			\midrule
			w/o Net & 0.0869 & 0.1343 & 0.1714 & 0.0553 & 0.0705 & 0.0803 \\ 
			ArcRec-REF & 0.0840 & 0.1268 & 0.1609 & 0.0555 & 0.0692 & 0.0782 \\ 
			\midrule
			w/o AWTP & 0.0811 & 0.1304 & 0.1687 & 0.0533 & 0.0691 & 0.0792 \\
			ArcRec-IA & 0.0721 & 0.1157 & 0.1495 & 0.0449 & 0.0589 & 0.0675 \\
			ArcRec-HA & 0.0595 & 0.1023 & 0.1414 & 0.0380 & 0.0518 & 0.0622 \\
			\midrule
			ArcRec & \textbf{0.0943} & \textbf{0.1442} & \textbf{0.1836} & \textbf{0.0614} & \textbf{0.0773} & \textbf{0.0875} \\ \bottomrule
	\end{tabular}
\end{table}

To demonstrate how different components of the proposed framework affect the performance, we conduct ablation studies by removing or replacing key modeling modules from the full architecture. Specifically, we focus on two key modules: the attributed reference network (ARN) and the attribute-level willingness to pay (AWTP) modeling. 

The ARNs serve as the core mechanism in developing our methodological framework, which greatly differ from prior work in utilizing attribute information and assist in better reference points modeling. In particular, we develop several variants of the proposed method to validate the effectiveness of this modeling component. Specifically, we remove the ARNs and the corresponding GNNs to obtain the model variant \textsf{w/o Net}, in which we directly use the initialized node embeddings as input for subsequent utility modeling. Meanwhile, we forego the ARNs and instead directly adopt the raw reference network without considering the decomposition as input for the GNNs layer, denoted as \textsf{ArcRec-REF}. This variant allows us to evaluate the effectiveness of decomposing the product reference relations based on attributes.

On the other hand, the proposed AWTP module captures consumers' personalized salience towards specific attributes in a data-driven manner, serving as a specially designed attention network. This provides another novel modeling perspective over classical models. Therefore, in the model variant \textsf{w/o AWTP}, we remove the AWTP module and assign equal weights to different attributes. We also implement the AWTP module in Equation \eqref{eq:demand_final} through a feedforward neural network with  consumer and attribute IDs as input, resulting in a model variant \textsf{ArcRec-IA}. Finally, in the model variant \textsf{ArcRec-HA}, the AWTP module is implemented by a feedforward neural network with consumers' purchased products along with consumer and attribute IDs as input to determine the attention weights. The ablation results are shown in Table \ref{tab:ablation}. 

\textbf{Benefits of attributed reference networks.} As summarized in Table \ref{tab:ablation}, the full model outperforms \textsf{w/o Net} for about 9\% and 11\% in terms of $\text{HR}@5$ and $\text{nDCG}@5$ respectively, which demonstrate the values in introducing reference network as well as ARNs. It also validates the assumption that the ``wisdom of crowds'' can contribute to more nuanced modeling of consumer's reference points than individual consumer’s behaviors alone. Moreover, it is seen that \textsf{ArcRec-REF} is also inferior to our proposed method. This can be attributed to the fact that the raw reference network without considering the decomposition with specific attributes may arouse invalid transitivity and thereby fail to capture appropriate reference points as well as fine-grained attribute-level utilities. Conversely, the ARNs in our proposed method can mitigate the gap and achieve better performances.

\textbf{Benefits of attribute-level willingness to pay.} 
The AWTP module in ArcRec plays a crucial role in capturing consumers' attitudes towards different attributes. By comparing the full ArcRec model with the variant \textsf{w/o AWTP} in Table \ref{tab:ablation}, we observe that the proposed AWTP module significantly improves the recommendation performance, which partly supports the necessity of assigning personalized weights for different attributes in modeling consumers' purchase decisions.
However, it is worth noting that the weights of AWTP cannot be effectively captured by traditional attention network designs. This is evident from the inferior performance of the variants \textsf{ArcRec-IA} and \textsf{ArcRec-HA} compared to the base model \textsf{w/o AWTP}, as shown in Table \ref{tab:ablation}. This further highlights the advantages of the construction of the AWTP based on the decomposed product embeddings and price deviations.

\subsubsection{A Post-hoc Analysis of Consumers' AWTP.}
The proposed method offers valuable insights into consumers' AWTP through the indicator $I_{u,k}$, as calculated by Equation \eqref{eq:attr_im}. This indicator quantifies the relative importance that consumers assign to different attributes when making purchasing decisions. By analyzing the values of $I_{u,k}$, we can gain a better understanding of how consumers weigh and prioritize different attributes. 

\begin{figure*}[t]
	\centering    
 \subfigure[]{\label{fig:tsne}\includegraphics[width=0.42\textwidth]{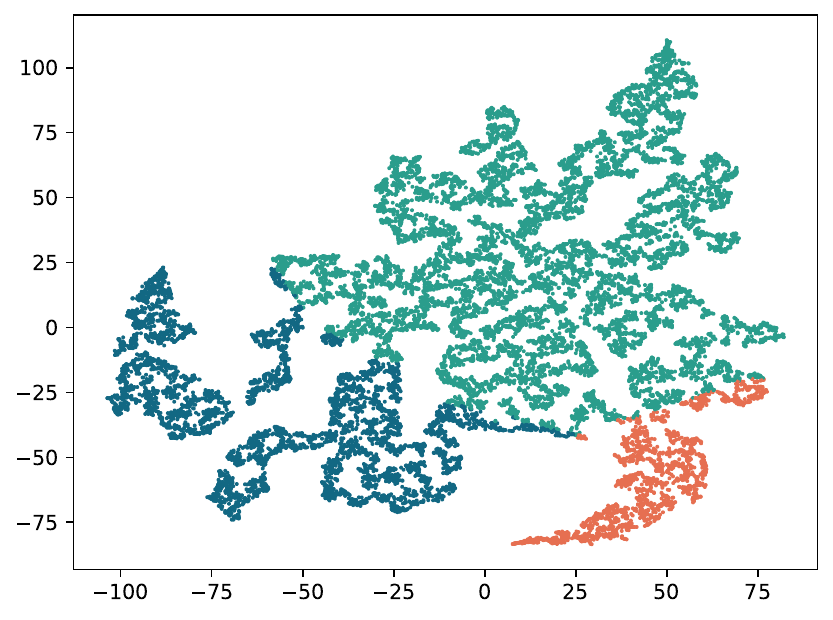}}
    \hspace{1cm}
    \subfigure[]
    {\label{fig:user_case}\includegraphics[width=0.4\textwidth]{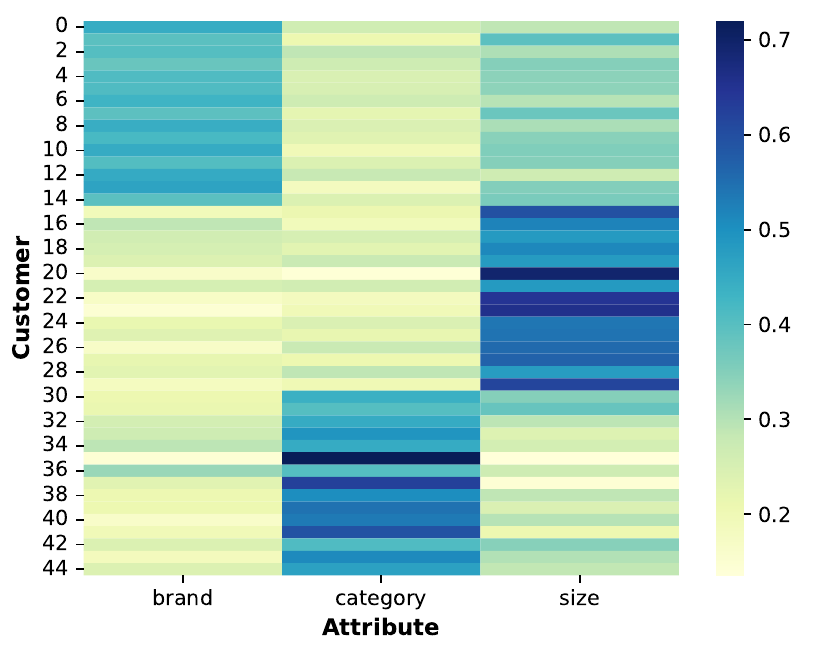}}
	\caption{Visualization of consumers' attribute-level willingness to pay (AWTP). (a) consumer segmentation based on AWTP. (b) Heat map of sampled consumers' AWTP.}
	\label{fig:case_study}
\end{figure*}

Then, to visualize the representation of the learned AWTP, we use the t-SNE method \citep{van2008visualizing} to project the indicator vector $I_{u,k}$ into a two-dimensional space. In addition, we further segment all the consumers into several clusters using K-Means clustering method based on the AWTP indicators, opting for 3 clusters. Each cluster represents a group of consumers with similar AWTP. Figure \ref{fig:tsne} presents a two-dimensional visualization of consumer distribution generated by the t-SNE method. Each dot in the figure represents an individual consumer, and the color of the dot indicates the cluster membership determined by the K-Means clustering method. As can be seen from the figure, different clusters are clearly separated from each other. This clustering visualization result demonstrates the effectiveness of the learned AWTP as a valuable tool for consumer segmentation.

Additionally, we also present a heat map to provide more detailed information of consumers' AWTP in Figure \ref{fig:user_case}. In particular, we randomly sample 15 consumers from each cluster, resulting in a total of 45 sampled consumers for the sake of visualization. The x-axis represents three different product attributes and the y-axis represents the sampled 45 consumers. The color of each grid in the heat map represents the value of the AWTP score for a specific pair of attribute and consumer. Darker colors indicate higher AWTP scores, indicating that the corresponding attribute is more important for the consumer. 

By examining the heat map, we can observe distinct patterns for each cluster. Consumers in the first cluster exhibit a higher AWTP for the ``brand" attribute, as indicated by the darker colors in the corresponding grid cells. On the other hand, consumers in the second cluster show a higher AWTP for the ``category" attribute, while consumers in the third cluster demonstrate a stronger priority for the ``size" attribute.

In Figure \ref{fig:case_study}, we present several representative cases to further validate the effectiveness of the AWTP and its ability to capture consumers' preferences. We randomly select one consumer from each cluster and display her AWTP scores for different attributes, along with her historical purchasing records. As indicated by the high AWTP score in the "brand" attribute, we can observe that consumer 10775 shows a strong preference for well-known brands such as \emph{Lancome}, \emph{Dior}, regardless of the price, category, or size of the products they purchase. Another representative consumer is 1122 with the highest AWTP score in ``category'', and we can also observe that the consumer's purchasing records are solely focused on the category of perfume. This suggests that the consumer may consider the store as a professional perfume store, and their purchasing decisions are greatly influenced by the specific category. 
In addition, we also find that consumer 5876 prefers to try sample products with small sizes, which indicates that the consumer may be more interested in sample packs or small-sized products. This aligns with the highest AWTP in the "size" attribute. 

In conclusion, the AWTP, learned by our proposed model, serves as an effective representation of consumers' attitude to different attributes. It enables us to uncover fine-grained preferences of consumers based on limited purchase behaviors, and thus can be utilized as a powerful tool for consumer segmentation and marketing strategy design. Moreover, the learned AWTP provides insights into consumers' priorities for different product attributes, which allows for a more detailed understanding of consumer preferences beyond overall product choices.

\begin{figure}[t]
	\centering
	\includegraphics[width=0.995\textwidth]{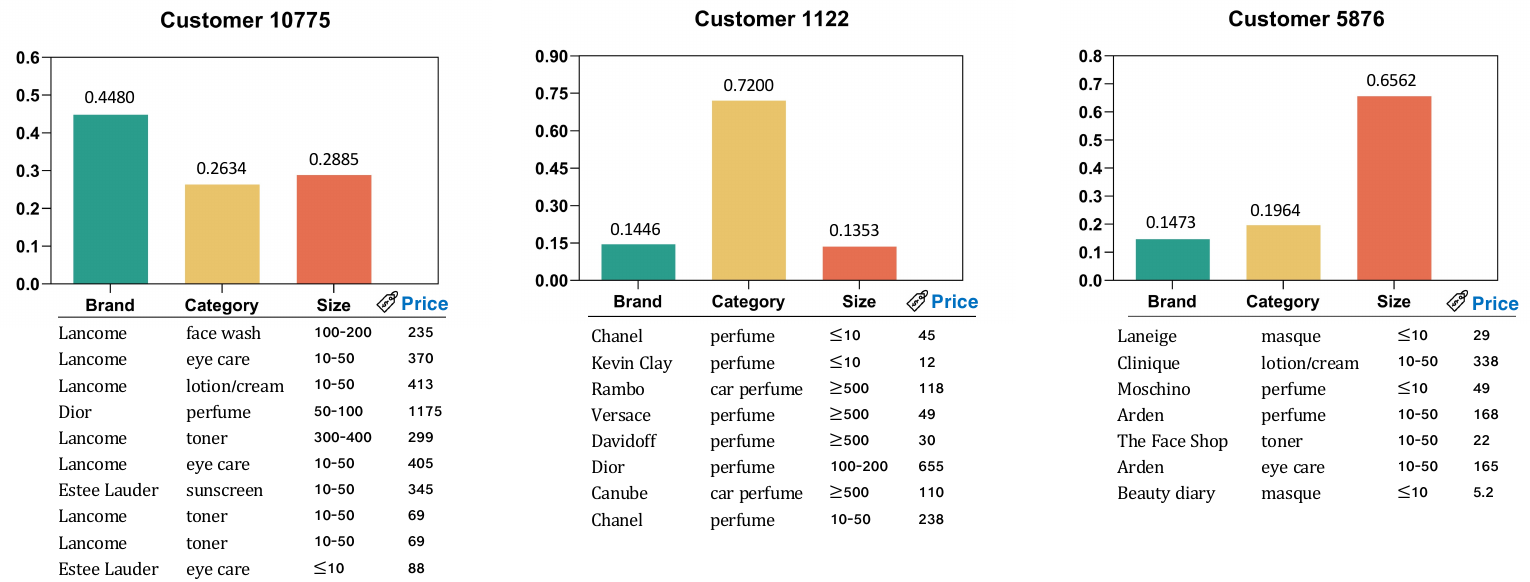}
	\caption{Representative consumers' AWTP and purchasing records.}
	\label{fig:case_study}
\end{figure}

\subsubsection{Parameter Analysis.} 
The number of GNN layers $L$ and the embedding dimension $d$ are critical hyperparameters of our model. In this experiment, we investigate the effects of these hyperparameters on ArcRec using the real-world dataset. The evaluation metrics employed are $\text{HR}@K$ and $\text{nDCG}@K$.

\begin{figure*}[t]
\centering    
    \subfigure[]{\label{fig:layer}\includegraphics[width=0.65\textwidth]{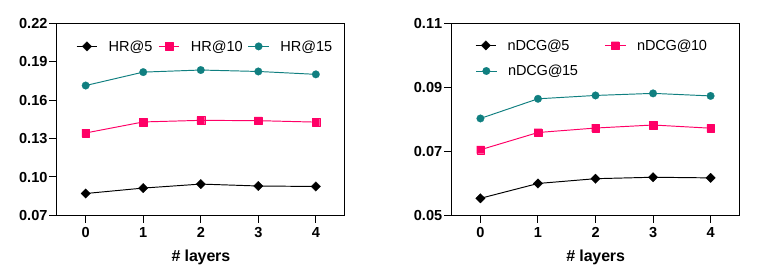}}
    \subfigure[]{\label{fig:emb}\includegraphics[width=0.65\textwidth]{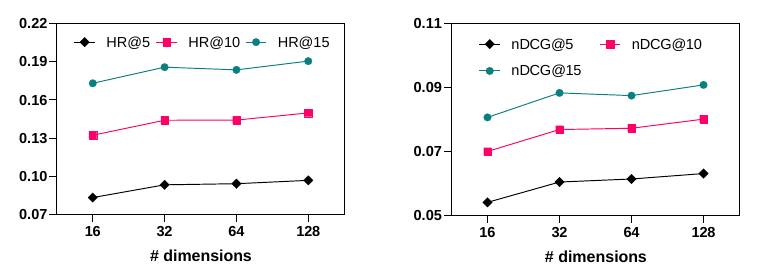}}
\caption{Impact of different hyperparameters on model performances. (a) $L$: the number of GNN layers. (b) $d$: embedding dimension.}
\label{fig:parameter_analysis}
\end{figure*}

\textbf{The number of GNN layers $L$.} As can be seen from Figure~\ref{fig:layer}, the model performance initially improves as $L$ increases and then tends to stabilize after $L$ exceeds 2. Notably, when $L=0$, our model is simplified to the model variant w/o Net proposed in the ablation study. As evident in Figure~\ref{fig:layer}, it's clear that the model performance is significantly lower when $L=0$ compared to cases with larger $L$, regardless of the specific value of $L$. This trend conforms to the performances observed in the ablation study and again highlights the necessity of introducing reference network as well as ARNs. 

\textbf{The embedding dimension $d$.} Figure \ref{fig:emb} illustrates the influence of the embedding dimension on the model performance. It's evident that, in most cases, model performance increases with the embedding dimension because larger embedding dimensions tend to provide stronger representational power. Nevertheless, employing high-dimensional representations doesn't consistently yield superior results. For instance, in Figure \ref{fig:emb}, we observe that both $\text{HR}@15$ and $\text{nDCG}@15$ metrics perform better when the embedding dimension is set to 32 compared to when it's set to 64.

\section{Discussions and Conclusions}
Choice modeling has become a longstanding problem in management science and marketing science fields, which is also the key module in improving recommendation effectiveness. Nowadays, numerous empirical studies have suggested that consumers' evaluations are reference-dependent and comparative in nature \citep{mussweiler2003comparison, kahneman2013prospect}, where consumers make decisions by comparing a target product to reference points. However, traditional recommender systems haven't explicitly integrated such theoretical foundation to effectively capture consumers' preferences. In order to pursue the benefits of better and realistic consumers' decision-making modeling, we propose to study the problem of exploring reference-dependent choice model from a data-driven perspective with the goal of developing an advanced recommendation model. Therefore, our study primarily makes contributions to the existing literature by formulating a new research problem.

Meanwhile, our study contributes to the existing information systems literature by advancing method development and introducing a novel method to the expanding repertoire of machine learning techniques designed to address significant business challenges. In our novel model, we combine deep neural network structure with the classical behavioral economic theories to derive an end-to-end learning framework. Building upon the idea of reference points in Prospect Theory, we successfully link the theory with recommendation methods by constructing the reference-dependent utility. Specially, to help reference points operationalization, we propose to integrate the ``wisdom of crowds'' and construct a reference network based on the co-purchasing records firstly. The reference network offers a global view of the collective purchasing behaviors, signifying the potential effects for reference point formation driven by some other factors such as the “also-viewed” or “also-purchased” recommendation links. To extend the reference-dependent choice to a more desirable attribute granularity, we further introduce the ARNs, which allow us to preserve appropriate product reference relations in the raw reference network for fine-grained reference points modeling.  The prevailing GNNs are applied to derive product representations from these networks, which can 
subsequently be utilized for interest-inspired and price-inspired preferences modeling to account for the two major gauges in consumer choices by designing an attribute-aware price sensitivity mechanism. Finally, to assemble the attribute-level reference-dependent utilities into an integrated one, we design a novel AWTP measurement which gains insights in understanding consumers' personalized trade-offs among different product attributes. Empirical evaluations in real-world data have demonstrated the effectiveness of the proposed method compared to the state-of-the-art recommendation methods, and the model also exhibits robustness against cold-start products. In addition, we also demonstrate that our model can well capture price effects through a controlled experiment, validating its ability to accurately respond to variations in pricing strategies.


Our proposed method has the following managerial implications.
Firstly, the newly designed recommender systems in this paper can provide a novel perspective in making recommendations that cater to consumers' fine-grained preferences and desired aspects for a product, which has the potential to revolutionize the way recommendations are made. In particular, by incorporating product price, the system is capable of delivering more cost-effective recommendations, recommending products that align with consumers' price perceptions. Additionally, the proposed recommender system can provide interpretations for its recommendations, adding transparency to guide consumers' decision process. For instance, by inferring the AWTP, the system can offer insights such as ``\emph{We recommend this product because you may prefer the brand}" or ``\emph{The brand is worth the price}". This level of interpretability can foster trust and confidence in the recommendations, enhancing the overall user experience. With this newly designed recommendation method, businesses can strengthen stronger relationships with their consumers, leading to repeat purchases, positive word-of-mouth, and long-term loyalty. This, in turn, can boost consumer lifetime value and drive sustained business growth. 

Second, since the method can effectively capture consumers' attitudes toward different attributes describing the products, partially reflected by the AWTP, it has the potential to provide nuanced guidance in devising promotional strategies for launching effective marketing campaigns. In particular, the clear separation of the consumer segments from the AWTP indicates that consumers within each segment share similar attribute preferences. This, in turn, empowers businesses to customize their marketing communications and promotional initiatives, spotlighting the attributes that resonate most with their target consumers, leading to more effective and targeted marketing campaigns. Moreover, the AWTP provides valuable insights into how consumers perceive and value different attributes in relation to price, which can also be leveraged by businesses to develop targeted pricing strategies. For example, a phone company may identify a consumer group highly values the storage capacity of smartphones through measuring AWTP, which could guide the company in justifying a premium price for models with extra storage. A pertinent real-world illustration of this strategy can be found in Apple's pricing approach with its iPhones. Apple offers variations of its iPhone models with differing storage capacities, pricing models with larger storage options at a premium. 

Third, the proposed method, with its ability to estimate consumers' AWTP from past purchases, can be a valuable tool in guiding product design, development, and the creation of new product lines. By understanding consumers' fine-grained preferences with respect to product attributes, companies can make more informed decisions about product attributes that are likely to resonate with their target audience. For instance, a phone company might introduce a new smartphone model based on AWTP findings, indicating that customers place a premium on features like extended battery life or more advanced camera capabilities. Therefore, this data-driven approach allows for a more tailored product development process, ensuring that new products can satisfy market demands and consumer preferences.

\theendnotes

\bibliographystyle{apalike}
\bibliography{biblio}









\end{document}